\documentclass{article}

\usepackage{microtype}
\usepackage{graphicx}
\usepackage{subfigure}
\usepackage{booktabs} 
\usepackage[T1]{fontenc}

\usepackage{hyperref}

\usepackage[accepted]{icml2025}

\usepackage{amsmath}
\usepackage{amssymb}
\usepackage{mathtools}
\usepackage{amsthm}

\usepackage[capitalize,noabbrev]{cleveref}

\theoremstyle{plain}

\theoremstyle{definition}

\theoremstyle{remark}

\usepackage[textsize=tiny]{todonotes}

\hypersetup{
    colorlinks,
    linkcolor={red!50!black},
    urlcolor={blue!80!black}
}
\usepackage{bm}
\usepackage{dsfont}
\usepackage{multirow}
\usepackage{colortbl}
\usepackage{xcolor}

\definecolor{cred}{HTML}{FF6B6B}
\definecolor{cyellow}{HTML}{FEC260}
\definecolor{cgreen}{HTML}{70AD47}
\definecolor{cblue}{HTML}{4D96FF}
\definecolor{cpurple}{HTML}{2A0944}
\definecolor{ggray}{RGB}{127,127,127}
\definecolor{aliceblue}{rgb}{0.94, 0.97, 1.0}

\newcommand{\type}[1]{\color{gray}{\scriptsize{#1}}}
\newcommand{\myparagraph}[1]{\textbf{#1}\hspace{1.8ex}}

\renewcommand{\thefootnote}{\fnsymbol{footnote}}

\makeatletter
\newcommand{\ssymbol}[1]{$^{\@fnsymbol{#1}}$}
\makeatother

\icmltitlerunning{MoH: Multi-Head Attention as Mixture-of-Head Attention}

\begin{document}

\twocolumn[
\icmltitle{MoH: Multi-Head Attention as Mixture-of-Head Attention}



\icmlsetsymbol{equal}{*}

\begin{icmlauthorlist}
\icmlauthor{Peng Jin}{1,2,3}
\icmlauthor{Bo Zhu}{-}
\icmlauthor{Li Yuan}{1,2,3,+}
\icmlauthor{Shuicheng Yan}{4,-}
\end{icmlauthorlist}

\icmlaffiliation{1}{School of Electronic and Computer Engineering, Shenzhen Graduate School, Peking University, Shenzhen, China}
\icmlaffiliation{2}{Pengcheng Laboratory, Shenzhen, China}
\icmlaffiliation{3}{School of AI for Science, Shenzhen Graduate School, Peking University, Shenzhen, China}
\icmlaffiliation{4}{National University of Singapore, Singapore}
\icmlaffiliation{+}{Rabbitpre Intelligence, Shenzhen, China}
\icmlaffiliation{-}{Skywork AI, Singapore}

\icmlcorrespondingauthor{Li Yuan}{yuanli-ece@pku.edu.cn}
\icmlcorrespondingauthor{Shuicheng Yan}{shuicheng.yan@gmail.com}

\icmlkeywords{Machine Learning, ICML}

\vskip 0.3in
]



\printAffiliationsAndNotice{}  

\begin{abstract}
In this work, we upgrade the multi-head attention mechanism, the core of the Transformer model, to reduce computational costs while maintaining or surpassing the previous accuracy level. We show that multi-head attention can be expressed in the summation form. Drawing on the insight that not all attention heads hold equal significance, we propose Mixture-of-Head attention (MoH), a new architecture that treats attention heads as experts in the Mixture-of-Experts (MoE) mechanism. MoH has two significant advantages: First, MoH enables each token to select the appropriate attention heads, enhancing inference efficiency without compromising accuracy or increasing the number of parameters. Second, MoH replaces the standard summation in multi-head attention with a weighted summation, introducing flexibility to the attention mechanism and unlocking extra performance potential. Extensive experiments on ViT, DiT, and LLMs demonstrate that MoH outperforms multi-head attention by using only 50\%$\sim$90\% of the attention heads. Moreover, we demonstrate that pre-trained multi-head attention models, such as LLaMA3-8B, can be further continue-tuned into our MoH models. Notably, MoH-LLaMA3-8B achieves an average accuracy of 64.0\% across 14 benchmarks, outperforming LLaMA3-8B by 2.4\% by utilizing only 75\% of the attention heads. We believe the proposed MoH is a promising alternative to multi-head attention and provides a strong foundation for developing advanced and efficient attention-based models. The code is available at \href{https://github.com/SkyworkAI/MoH}{https://github.com/SkyworkAI/MoH}.
\end{abstract}

\section{Introduction}
Since attention is introduced and becomes a fundamental component of Transformers~\citep{vaswani2017attention}, multi-head attention has been the standard architecture for natural language processing~\citep{kenton2019bert} and computer vision tasks~\citep{dosovitskiy2021an}. It is well known that using multiple heads can improve model accuracy. However, not all attention heads hold equal significance. Some works have shown that many attention heads can be pruned without affecting accuracy. For example, \citet{voita2019analyzing} introduces a method to quantify the usefulness of each attention head and prune those that are redundant. Similarly, \citet{michel2019sixteen} challenges the necessity of multiple heads by examining the impact of extensive pruning across various settings. In computer vision, some works also identify attention head redundancy. \citet{bhattacharyya2023decatt} reduces redundancy to boost performance, while \citet{yun2024shvit} develop single-head attention for efficiency. These findings demonstrate that vanilla multi-head attention contains redundant attention heads.

\begin{figure*}[tbp]
\centering
\includegraphics[width=1\textwidth]{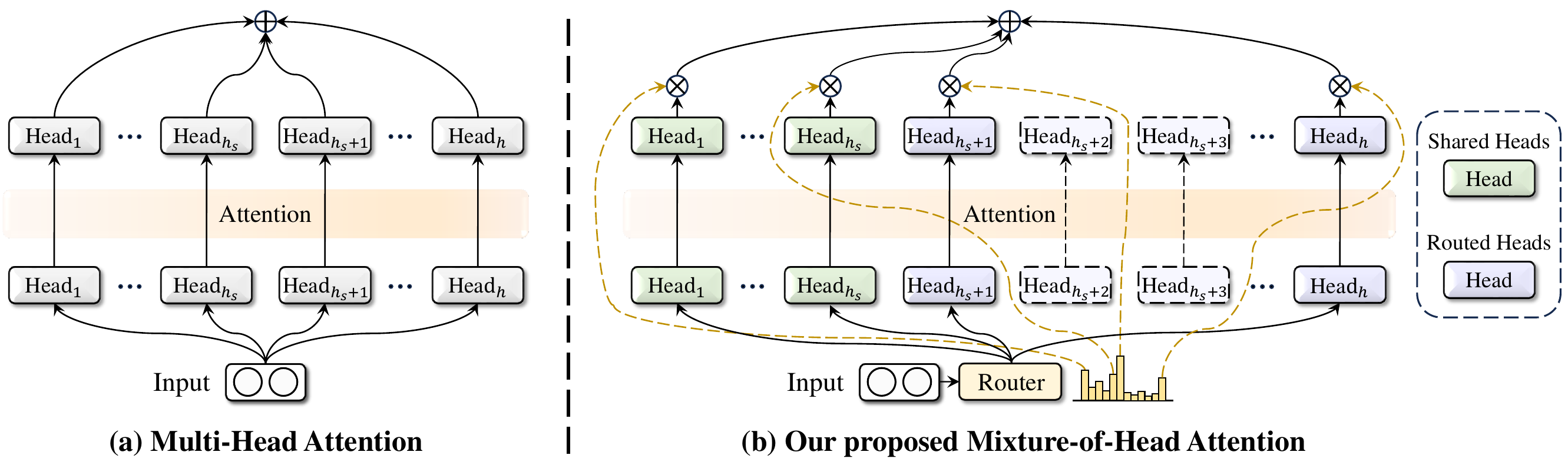}
\vspace{-2.em}
\caption{\textbf{A high-level comparison between the multi-head attention and our proposed mixture-of-head attention.} Subfigure (a) illustrates a standard multi-head attention layer with $h$ attention heads, while subfigure~(b) demonstrates our proposed Mixture-of-Head attention (MoH) architecture. It is important to note that MoH does not increase the number of attention heads, ensuring that the total parameter for MoH is comparable to that of the multi-head attention.}
\vspace{-.5em}
\label{fig1}
\end{figure*}

Besides, in multi-head attention, each head operates in parallel, and the final output is the sum of all heads (please refer to Section~\ref{method mha}). Given that these attention heads operate independently and some may be redundant, we argue that it is possible to build a dynamic attention-head routing mechanism. Such a mechanism would enable each token to adaptively select the appropriate attention heads, enhancing inference efficiency without compromising accuracy.

To this end, we introduce Mixture-of-Head attention (MoH), a new architecture that integrates multi-head attention with the Mixture-of-Experts (MoE) mechanism~\citep{jacobs1991adaptive}. Specifically, we propose to treat attention heads as experts within the MoE framework. Similar to MoE, MoH consists of multiple attention heads and a router that activates the Top-K heads for each token. Moreover, we replace the standard summation in multi-head attention with a weighted summation. This design offers two significant advantages: \textbf{First}, MoH allows each token to select the most relevant attention heads, improving inference efficiency without sacrificing accuracy or increasing the parameters. \textbf{Second}, by replacing the standard summation in multi-head attention with a weighted summation, MoH enhances the flexibility of the attention mechanism and increases the performance potential. Moreover, to efficiently capture common knowledge across different contexts, we designate a subset of attention heads as shared heads that remain always activated.

We evaluate our proposed MoH across various popular model frameworks, including Vision Transformers~(ViT)~\citep{dosovitskiy2021an} for image classification, Diffusion models with Transformers~(DiT)~\citep{peebles2023scalable} for class-conditional image generation, and Large Language Models~(LLMs)~\citep{brown2020language,chatgpt,ouyang2022training}. We show that MoH achieves competitive performance, or even outperforms multi-head attention with only 50\%$\sim$90\% of the attention heads. For example, MoH-ViT-B achieves 84.9\%/84.7\% Top-1 accuracy on the ImageNet-1K~\citep{deng2009imagenet} classification benchmark, surpassing well-tuned multi-head attention baselines with only 75\%/50\% of the attention heads.

Furthermore, we demonstrate that pre-trained multi-head attention models, such as LLaMA3-8B~\citep{dubey2024llama}, can be further continue-tuned into our MoH models. Specifically, using only about 3\%~(400B tokens) of the original LLaMA3 pre-training data for continue-tuning, MoH-LLaMA3-8B achieves an average accuracy of 64.0\% across 14 benchmarks, outperforming LLaMA3-8B by 2.4\% by utilizing only 75\% of the attention heads. These results show that MoH is a promising alternative to vanilla multi-head attention, laying a solid foundation for developing advanced and efficient attention-based models. The main contributions are summarized as follows:

\begin{itemize}
    \item We propose a dynamic attention-head routing mechanism that allows each token to adaptively select the appropriate attention heads, enhancing model performance and inference efficiency without increasing the number of parameters.
    \item In addition to training from scratch, we demonstrate that pre-trained multi-head attention models, such as LLaMA3-8B, can be further continue-tuned into our MoH models, greatly enhancing the applicability of the proposed MoH method.
    \item Extensive experiments across various popular model frameworks, including ViT, DiT, and LLMs, confirm that MoH is a promising alternative to vanilla multi-head attention, laying a solid foundation for developing advanced and efficient attention-based models.
\end{itemize}

\section{Related Work}
\myparagraph{Multi-Head Attention.} Transformers~\citep{vaswani2017attention} have garnered significant interest and success in both natural language processing and computer vision. The success of transformers has been long attributed to the multi-head attention mechanism~\citep{cordonnier2020multi}. Multi-head attention mechanism is proposed by \citet{vaswani2017attention} to enhance the representation power of an attention layer by allowing multiple attention heads to operate on different low-dimensional projections of the input. The outputs from these heads are then concatenated to form the final result. Alternatively, by decomposing the output projection matrix by rows, multi-head attention can be expressed in a summation form. In summation form, each head operates in parallel, and the final output is the sum of all heads. Inspired by this observation, we propose MoH, a dynamic attention-head routing mechanism that allows each token to adaptively select the appropriate heads.

\myparagraph{Mixture-of-Experts Models.} 
The Mixture-of-Experts (MoE) method~\citep{du2022glam,lewis2021base,rajbhandari2022deepspeed,roller2021hash,zhou2022mixture,jin2024moe++} is introduced to expand the capacity of deep neural networks without increasing computational costs. In this approach, only a subset of parameters, known as experts, is activated for each input. \citet{shazeer2017outrageously} first introduces an MoE layer between LSTM layers. Switch Transformer~\citep{fedus2022switch} further simplifies the gating mechanism by selecting only the Top-1 expert per token. Gshard~\citep{lepikhin2020gshard} improves the Top-2 expert routing strategy. In contrast to MoE, which emphasizes efficient parameter scaling while maintaining manageable computational costs, our MoH focuses on reducing the activation of redundant attention heads without increasing the number of parameters.

\myparagraph{Attention Head Specialization and Efficiency.} Many recent studies show that not all attention heads in Transformers are equally useful. \citet{peng2020mixture} proposes a mixture-of-heads approach, where only a few selected heads are used, yet the model performs just as well or even better. \citet{csordas2024switchhead} pushes this further with SwitchHead, an MoE-style method that activates only a small number of heads for each token, speeding up inference while keeping performance high. In long-context language models, this idea is even more clear. \citet{wu2024retrieval} shows that a few special retrieval heads are mainly responsible for keeping facts consistent in long inputs. \citet{fu2024not} finds that keeping only the most useful heads in the KV cache can save memory. \citet{xiao2024duoattention} proposes DuoAttention, which combines different types of heads to make long-context inference more efficient without losing quality. Similar patterns appear in vision models. \citet{gandelsman2023interpreting} shows that CLIP's attention heads each focus on specific visual features, and this can be explained through related text prompts. \citet{balasubramanian2024decomposing} finds that this kind of head specialization also exists in other vision models beyond CLIP. \citet{basile2024residual} shows that using only a few selected heads chosen by spectral methods can even beat the full model on zero-shot tasks.

\section{Methodology}
In this work, we aim to reduce the activation of redundant attention heads without increasing the number of parameters. A high-level comparison between the vanilla multi-head attention and our proposed MoH is presented in Fig.~\ref{fig1}.

\subsection{Multi-Head Attention}\label{method mha}
We begin by reviewing the multi-head attention mechanism introduced by \citet{vaswani2017attention}. The multi-head attention mechanism is based on scaled dot-product attention. Specifically, for $T$ tokens $\bm{X} \in \mathbb{R}^{T \times d_{in}}$ of $d_{in}$ dimensions each and $T'$ tokens $\bm{X}' \in \mathbb{R}^{T' \times d_{in}}$ of $d_{in}$ dimensions each, the scaled dot-product attention is computed as follows:
\begin{equation}
\begin{aligned}
\textrm{Attention}(\bm{Q},\bm{K}&,\bm{V})=\textrm{Softmax}\Big(\frac{\bm{Q}\bm{K}^\top}{\sqrt{d_k}} \Big)\bm{V}, \\
\bm{Q}=\bm{X}\bm{W}_Q,\ &\bm{K}=\bm{X}'\bm{W}_K,\ \bm{V}=\bm{X}'\bm{W}_V,
\end{aligned}
\end{equation}
where $\bm{W}_Q \in \mathbb{R}^{d_{in} \times d_{k}}$, $\bm{W}_K \in \mathbb{R}^{d_{in} \times d_{k}}$, and $\bm{W}_V \in \mathbb{R}^{d_{in} \times d_{v}}$ represent the projection matrices for the query, key, and value, respectively. In self-attention, the input tokens are the same, i.e., $\bm{X}'=\bm{X}$, and it is common for the key and value dimensions to be equal, i.e., $d_{v}=d_{k}$.

\myparagraph{Concatenation Form.} 
To enhance the representation power, \citet{vaswani2017attention} proposes to allow multiple attention heads to operate on different low-dimensional projections of the input tokens. Specifically, the multi-head attention mechanism computes $h$ different low-dimensional projections of $(\bm{Q}, \bm{K}, \bm{V})$, performs scaled dot-product attention for each head, concatenates the results, and applies a projection to the concatenated output. The concatenation form of the multi-head attention can be formulated as:
\begin{equation}
\begin{aligned}
\textrm{MultiHead}(&\bm{X},\bm{X}')=\textrm{Concat}(\bm{H}^{1}, \bm{H}^{2}, ..., \bm{H}^{h})\bm{W}_O, \\
\bm{H}^{i}=&\textrm{Attention}(\bm{X}\bm{W}_Q^{i},\bm{X}'\bm{W}_K^{i},\bm{X}'\bm{W}_V^{i}),
\end{aligned}
\end{equation}
where $\bm{W}_{Q}^{i} \in \mathbb{R}^{d_{in} \times d_{k}/h}$, $\bm{W}_{K}^{i} \in \mathbb{R}^{d_{in} \times d_{k}/h}$, and $\bm{W}_{V}^{i} \in \mathbb{R}^{d_{in} \times d_{v}/h}$ represent the $i_{th}$ projection matrices for the query, key, and value, respectively. $\bm{W}_O \in \mathbb{R}^{d_{v} \times d_{out}}$ is the final output projection matrix.

\myparagraph{Summation Form.}
The multi-head attention mechanism is typically represented in its concatenation form. However, from another perspective, if we decompose $\bm{W}_O \in \mathbb{R}^{d_{v} \times d_{out}}$ by rows, we can express multi-head attention in a summation form. Specifically, $\bm{W}_O$ can be divided into $h$ matrices by rows, i.e., $[\bm{W}_{O}^{1}, \bm{W}_{O}^{2}, ..., \bm{W}_{O}^{h}]=\bm{W}_O$, where $\bm{W}_{O}^{i} \in \mathbb{R}^{d_{v}/h \times d_{out}}$. Finally, the summation form of the multi-head attention can then be formulated as:
\begin{equation}
\textrm{MultiHead}(\bm{X},\bm{X}')=\sum_{i=1}^{h} \bm{H}^{i}\bm{W}_O^{i}.
\label{multi-head attention}
\end{equation}
The concatenation form can be viewed as a variant of the summation form, where the sum of the dimensions of all attention heads is exactly equal to the hidden size. As shown in Eq.~\ref{multi-head attention}, in standard multi-head attention, each attention head operates in parallel, and the final output is the sum of all attention heads. Since these attention heads function independently, we can build a dynamic attention-head routing mechanism allowing each token to adaptively select the most relevant attention heads, improving inference efficiency without compromising accuracy.

\begin{table*}[t]
\footnotesize
\centering
\caption{\textbf{Comparisons to current state-of-the-art methods on ImageNet-1K classification.} Our MoH-ViT models, based on TransNeXt~\citep{shi2024transnext}, are trained for 300 epochs using a resolution of 224$\times$224. To ensure a fair comparison, we only replace the standard multi-head attention with our Mixture-of-Head attention~(MoH), keeping all other training parameters identical to TransNeXt.}
\vspace{.5em}
\begin{minipage}[c]{0.49\textwidth}
\resizebox{1.\linewidth}{!}
{
\begin{tabular}{lccc}
\toprule[1.25pt]
 \multirow{2}{*}{\textbf{Methods}} & \textbf{\#Params} & \textbf{\#Activated}  & \textbf{Acc} \\ 
 & \textbf{(M)} & \textbf{\ Heads (\%)} & \textbf{(\%)} \\ 
 \midrule
 DeiT-S~\tiny{\citep{touvron2021training}} & 22 & 100 & 79.8  \\
 T2T-ViT-19~\tiny{\citep{yuan2021tokens}} & 39 & 100 & 81.9 \\
 Swin-S~\tiny{\citep{liu2021swin}} & 50 & 100 & 83.1 \\
 PVTv2-B3~\tiny{\citep{wang2022pvt}} & 45 & 100  & 83.2 \\
 CoAtNet-1~\tiny{\citep{dai2021coatnet}} & 42 & 100  & 83.3 \\
 Focal-S~\tiny{\citep{yang2021focal}} & 51 & 100  & 83.5 \\
 FocalNet-S~\tiny{\citep{yang2022focal}} & 50 & 100  & 83.5 \\
 MViTv2-S~\tiny{\citep{li2022mvitv2}} & 35 & 100 & 83.6 \\
 UniFormer-B~\tiny{\citep{li2023uniformer}} & 50 & 100 & 83.9 \\
 CAFormer-S36~\tiny{\citep{yu2023metaformer}} & 39 & 100 & 84.5 \\
 TransNeXt-S~\tiny{\citep{shi2024transnext}} & 50 & 100 & \textbf{84.7}  \\
 \midrule
 \rowcolor{aliceblue!60} \textbf{MoH-ViT-S} & 50 & 80 & \textbf{84.7} \\
 \textbf{MoH-ViT-S} & 50 & 75 & 84.6 \\
\bottomrule[1.25pt]
\end{tabular}
}
\end{minipage}
\hfill \
\begin{minipage}[c]{0.48\textwidth}
\resizebox{1.\linewidth}{!}
{
\footnotesize
\begin{tabular}{lccc}
\toprule[1.25pt]
 \multirow{2}{*}{\textbf{Methods}} & \textbf{\#Params} & \textbf{\#Activated} & \textbf{Acc} \\ 
 & \textbf{(M)} & \textbf{\ Heads (\%)} & \textbf{(\%)} \\ 
 \midrule
 DeiT-B~\tiny{\citep{touvron2021training}} & 86 & 100 & 81.8  \\
 T2T-ViT-24~\tiny{\citep{yuan2021tokens}} & 64 & 100 & 82.3 \\
 Swin-B~\tiny{\citep{liu2021swin}} & 88 & 100 & 83.5 \\
 PVTv2-B5~\tiny{\citep{wang2022pvt}} & 82 & 100 & 83.8 \\
 Focal-B~\tiny{\citep{yang2021focal}} & 90 & 100 & 83.8 \\
 FocalNet-B~\tiny{\citep{yang2022focal}} & 89 & 100 & 83.9 \\
 CoAtNet-2~\tiny{\citep{dai2021coatnet}} & 75 & 100 & 84.1 \\
 MViTv2-B~\tiny{\citep{li2022mvitv2}} & 52 & 100 & 84.4 \\
 MOAT-2~\tiny{\citep{yang2022moat}} & 73 & 100 & 84.7 \\
 iFormer-L~\tiny{\citep{si2022inception}} & 87 & 100 & 84.8 \\
 TransNeXt-B~\tiny{\citep{shi2024transnext}} & 90 & 100 & 84.8  \\
 \midrule
 \rowcolor{aliceblue!60} \textbf{MoH-ViT-B} & 90 & 75 & \textbf{84.9} \\
 \textbf{MoH-ViT-B} & 90 & 50 & 84.7 \\
\bottomrule[1.25pt]
\end{tabular}
}
\end{minipage}
\label{tab: vit}
\vspace{-.5em}
\end{table*}

\subsection{Mixture-of-Head Attention}
Recently, the Mixture-of-Experts (MoE) method has emerged as a popular approach for scaling the parameters of large language models~\citep{jiang2024mixtral,muennighoff2024olmoe}. A MoE layer consists of multiple expert networks and a router that activates the Top-K experts. Generally, the number of activated experts $K$ is significantly smaller than the total number of experts to ensure inference efficiency.

\myparagraph{Heads as Experts.} 
Inspired by the great success of MoE, we propose Mixture-of-Head attention~(MoH), which treats attention heads as experts. Specifically, MoH consists of $h$ heads $\bm{H}=\{H^1, H^2, ..., H^h\}$ and a router that activates the Top-K attention heads. Formally, given input tokens $\bm{X}$ and $\bm{X}'$, the output of MoH is the weighted sum of outputs from the $K$ selected attention heads:
\begin{equation}
\textrm{MoH}(\bm{X},\bm{X}')=\sum_{i=1}^{h}g_{i} \bm{H}^{i}\bm{W}_O^{i},
\end{equation}
where $g_{i}$ represents the routing score. $g_{i}$ is non-zero only when the $i_{th}$ attention head is activated. This design provides two key advantages: (i)~On the one hand, MoH enables each token to select the most relevant attention heads, boosting inference efficiency while maintaining accuracy. (ii)~On the other hand, in contrast to the standard summation in multi-head attention, the weighted summation in MoH enhances the flexibility of the attention mechanism and unlocks performance potential.

\myparagraph{Shared Heads.}
In attention mechanism, some attention heads may capture common knowledge across different contexts, such as grammatical rules in language. Inspired by \citet{dai2024deepseekmoe}, we designate a subset of heads as shared heads that remain always activated. By consolidating common knowledge within shared heads, we reduce redundancy among the other dynamically routed heads.

\myparagraph{Two-Stage Routing.}
Moreover, to dynamically balance the weights between shared and routed heads, we propose a two-stage routing strategy. In this routing strategy, the routing scores are determined by both the score of each individual head and the score associated with the head type. Specifically, given the $t_{th}$ input token $\bm{x}_{t} \in \mathbb{R}^{d_{in}}$ in $\bm{X} \in \mathbb{R}^{T \times d_{in}}$, the routing score $g_{i}$ is defined as:
\begin{equation}
\quad g_{i}=
\begin{cases}
\alpha_1\text{Softmax}(\bm{W}_s \bm{x}_{t})_i, &\text{if}\ 1\leq i\leq h_s,\\
\alpha_2\text{Softmax}(\bm{W}_r \bm{x}_{t})_{{i-h_{s}}}, &\text{if Head $i$ is activated},\\
0, & \text{otherwise},
\end{cases}
\label{router}
\end{equation}
where $h_s$ denotes the number of shared heads. $\bm{W}_s \in \mathbb{R}^{h_s \times d_{in}}$ and $\bm{W}_r \in \mathbb{R}^{(h-h_s) \times d_{in}}$ represent the projection matrices for the shared and routed heads, respectively. If $(\bm{W}_r \bm{x}_{t})_{{i-h_{s}}}\in \text{Top-K}\big(\{(\bm{W}_r \bm{x}_{t})_{{i-h_{s}}} |h_{s}+1\leq i\leq h\}\big)$, then the routed Head $i$ is activated. The coefficients $\alpha_1$ and $\alpha_2$ balance the contributions of the shared and routed heads, and are defined as:
\begin{equation}
[\alpha_1, \alpha_2]=\text{Softmax}(\bm{W}_h\bm{x}_{t}),
\label{alpha}
\end{equation}
where $\bm{W}_h \in \mathbb{R}^{2\times d_{in}}$ is the trainable projection matrix, and $d_{in}$ is the hidden size of $\bm{x}_{t}$.

\begin{table*}[t]
\footnotesize
\centering
\caption{\textbf{Comparisons to current state-of-the-art methods on the benchmarking of class-conditional image generation on ImageNet-1K at 256$\times$256 resolution.} ``$\uparrow$'' denotes that higher is better. ``$\downarrow$'' denotes that lower is better. ``cfg'' denotes the classifier-free diffusion guidance scale. ``{\color{gray}{400K}}'' denotes the training budget is 400K training steps.}
\vspace{.5em}
\resizebox{1.\linewidth}{!}
{
\begin{tabular}{lcccccccccc}
\toprule[1.25pt]
 {\textbf{Methods}}  & \textbf{\#Params (M)} & \textbf{\#Activated Heads (\%)} & {\textbf{FID$\downarrow$}} & {\textbf{sFID$\downarrow$}} & {\textbf{IS$\uparrow$}} & {\textbf{Precision$\uparrow$}} & {\textbf{Recall$\uparrow$}}\\
 \midrule
 DiT-S/2 \type{400K}~\tiny{\citep{peebles2023scalable}} & 33 & 100 & 68.40 & - & - & - & - \\
 \rowcolor{aliceblue!60} \textbf{MoH-DiT-S/2} \type{400K} & 33 & 90 & \textbf{67.25} & \textbf{12.15} & \textbf{20.52} & \textbf{0.37} & \textbf{0.58} \\
 \textbf{MoH-DiT-S/2} \type{400K} & 33 & 75 & 69.42 & 12.85 & 19.96 & 0.36 & 0.55 \\
 \midrule
 \midrule
 DiT-B/2 \type{400K}~\tiny{\citep{peebles2023scalable}} & 130 & 100 & 43.47 & - & - & - & - \\
 \rowcolor{aliceblue!60} \textbf{MoH-DiT-B/2} \type{400K} & 131 & 90 & \textbf{43.40} & \textbf{8.40} & \textbf{33.51} & \textbf{0.49} & \textbf{0.63} \\
 \textbf{MoH-DiT-B/2} \type{400K} & 131 & 75 & 43.61 & 8.48 & 33.43 & \textbf{0.49} & 0.62 \\
 \midrule
 \midrule
 DiT-L/2 \type{400K}~\tiny{\citep{peebles2023scalable}} & 458 & 100 & 23.33 & - & - & - & - \\
 \rowcolor{aliceblue!60} \textbf{MoH-DiT-L/2} \type{400K} & 459 & 90 & \textbf{23.17} & \textbf{6.16} & \textbf{58.92} & \textbf{0.61} & \textbf{0.63} \\
 \textbf{MoH-DiT-L/2} \type{400K} & 459 & 75 & 24.29 & 6.38 & 57.75 & 0.60 & \textbf{0.63} \\
 \midrule
 \midrule
 DiT-XL/2 \type{7,000K}~\tiny{\citep{peebles2023scalable}} & 675 & 100 & 9.62 & 6.85 & 121.50 & 0.67 & \textbf{0.67} \\
 DiT-XL/2 {\type{7,000K}} (cfg=1.25)  & 675 & 100 & 3.22 & 5.28 & 201.77 & 0.76 & 0.62 \\
 \textbf{MoH-DiT-XL/2} \type{2,000K} & 676 & 75 & 10.95 & 6.19 & 106.69 & 0.67 & 0.66 \\
 \textbf{MoH-DiT-XL/2} \type{2,000K} & 676 & 90 & 10.67 & 6.15 & 107.80 & 0.67 & 0.65 \\
 \textbf{MoH-DiT-XL/2} \type{7,000K} & 676 & 90 & 8.56 & 6.61 & 129.54 & 0.68 & \textbf{0.67} \\
 \rowcolor{aliceblue!60} \textbf{MoH-DiT-XL/2} {\type{7,000K}} (cfg=1.25) & 676 & 90 & \textbf{2.94} & \textbf{5.17} & \textbf{207.25} & \textbf{0.77} & 0.63 \\
\bottomrule[1.25pt]
\end{tabular}
}
\label{tab: dit2}
\vspace{-.5em}
\end{table*}

\myparagraph{Load Balance Loss} 
Directly training an MoE layer often causes the majority of tokens to be routed to a small number of experts, leaving the remaining experts insufficiently trained~\citep{shazeer2017outrageously}. To avoid the unbalanced load in the proposed MoH, following previous MoE methods~\citep{lepikhin2020gshard,wei2024skywork}, we apply a load balance loss. Specifically, for the $t_{th}$ input token $\bm{x}_{t} \in \mathbb{R}^{d_{in}}$ in $\bm{X} \in \mathbb{R}^{T \times d_{in}}$, the load balance loss $\mathcal{L}_{b}$ is formulated as:
\begin{equation}
\begin{aligned}
\mathcal{L}_{b}=&\sum_{i=h_{s}+1}^{h}P_i f_i,\ P_i=\frac{1}{T}\sum_{t=1}^{T}\text{Softmax}(\bm{W}_r \bm{x}_{t})_{{i-h_{s}}},\\
&f_i=\frac{1}{T}\sum_{t=1}^{T}\mathds{1}(\text{Token $\bm{x}_t$ selects Head $i$}),
\end{aligned}
\end{equation}
where $T$ denotes the number of tokens. $\mathds{1}(*)$ denotes the indicator function.

\myparagraph{Total Training Objective.} 
It is worth noting that the MoH is a general framework. Therefore, we evaluate our proposed MoH across various popular model frameworks, including Vision Transformers~(ViT), Diffusion models with Transformers~(DiT), and Large Language Models~(LLMs). Depending on the specific task, we require the task-specific loss. Finally, the total training loss is the weighted sum of the task-specific loss $\mathcal{L}_{task}$ and the load balance loss $\mathcal{L}_{b}$:
\begin{equation}
\mathcal{L} = \mathcal{L}_{task} + \beta \mathcal{L}_{b},
\end{equation}
where $\beta$ is the trade-off hyper-parameter to mitigate the risk of routing collapse. By default, the weight $\beta$ for the load balance loss is set to 0.01 for all tasks.

\section{Experiments}\label{Experiments}
\subsection{ViT for Image Classification}
\myparagraph{Model Settings.} For Vision Transformers~(ViT)~\citep{dosovitskiy2021an}, our MoH-ViT models are implemented based on the TransNeXt~\citep{shi2024transnext} framework and trained from scratch on the ImageNet-1K dataset~\citep{deng2009imagenet}, which contains over 1.2 million images in 1,000 categories. To ensure a fair comparison, we only replace the standard multi-head attention with the proposed MoH, while keeping all other training parameters identical to TransNeXt.

\myparagraph{Training Details.} Our MoH-ViT models are trained for 300 epochs using automatic mixed precision across 8 GPUs. We follow the training strategy of TransNeXt, which includes various data augmentation techniques, including Random Augmentation~\citep{cubuk2020randaugment}, Mixup~\citep{zhang2017mixup}, CutMix~\citep{yun2019cutmix}, and Random Erasing~\citep{zhong2020random}. We also apply Label Smoothing~\citep{szegedy2016rethinking} and DropPath~\citep{huang2016deep} to regularize our models. We optimize our models using AdamW optimizer~\citep{loshchilov2017decoupled} with a gradient clipping norm of 1.0 and a weight decay of 0.05. The initial learning rate is set to 1e-3, with a 5-epoch warm-up starting at 1e-6. A cosine learning rate scheduler~\citep{loshchilov2016sgdr} is employed to decay the learning rate. During training, images are randomly cropped to a size of 224$\times$224. It is worth noting that we do not use Exponential Moving Average~(EMA) weights.

\begin{table*}[t]
\footnotesize
\centering
\caption{\textbf{Comparisons between MoH-LLMs and vanilla LLMs.} ``{\color{gray}{100B}}''  denotes a training budget of 100 billion tokens, while ``{\color{gray}{200B}}'' denotes a budget of 200 billion tokens. We observe that larger models, e.g., MoH-LLM-B, generally perform worse than smaller models, e.g., MoH-LLM-S, on TruthfulQA, consistent with the findings reported by \citet{lin-etal-2022-truthfulqa}.}
\vspace{.5em}
\resizebox{1.\linewidth}{!}
{
\begin{tabular}{lcccccccccc}
\toprule[1.25pt]
 {\textbf{Methods}} & \textbf{\#Activated Heads (\%)} & \textbf{SciQ} & \textbf{PIQA} & \textbf{WinoGrande} & \textbf{OpenbookQA} & \textbf{LogiQA} & \textbf{TruthfulQA}  & {\textbf{Average}} \\ 
 \midrule
 LLM-S \type{100B} & 100 & 63.0 & \textbf{63.1} & 51.1 & 27.4 & \textbf{26.9} & 31.6 & 43.9 \\
 \textbf{MoH-LLM-S} \type{100B} & 75 & 64.7 & 62.0 & 50.6 & 28.8 & 26.4 & 35.2 & 44.6 \\
 \rowcolor{aliceblue!60} \textbf{MoH-LLM-S} \type{100B} & 50 & \textbf{67.0} & 62.2 & \textbf{51.5} & \textbf{29.2} & 26.7 & \textbf{35.6} & \textbf{45.4} \\
 \midrule
 LLM-B \type{100B} & 100 & 73.1 & \textbf{69.7} & 52.0 & \textbf{31.8} & \textbf{28.4} & 29.5 & 47.4 \\
 \rowcolor{aliceblue!60} \textbf{MoH-LLM-B} \type{100B} & 75 & 74.7 & 69.2 & \textbf{52.8} & 30.0 & 28.1 & 32.2 & \textbf{47.8} \\
 \textbf{MoH-LLM-B} \type{100B} & 50 & \textbf{75.2} & 67.0 & 52.0 & 29.0 & 26.9 & \textbf{32.8} & 47.2 \\
 \midrule
 LLM-B \type{200B} & 100 & 73.1 & \textbf{70.3} & 53.3 & \textbf{32.4} & 29.0 & 29.5 & 47.9 \\
 \rowcolor{aliceblue!60} \textbf{MoH-LLM-B} \type{200B} & 75 & \textbf{76.0} & 69.2 & 52.7 & 30.4 & \textbf{29.8} & 32.6 & \textbf{48.5} \\
 \textbf{MoH-LLM-B} \type{200B} & 50 & 75.6 & 66.9 & \textbf{53.5} & 29.4 & 26.7 & \textbf{32.7} & 47.5 \\
\bottomrule[1.25pt]
\end{tabular}
}
\label{tab: LLMs}
\vspace{-.5em}
\end{table*}

\begin{table*}[t]
\footnotesize
\centering
\caption{\textbf{Comparisons between MoH-LLaMA3-8B and LLaMA3-8B.} {Please refer to Tab.~\ref{apdx tab: LLaMA3-8B-stage1} in the Appendix for the performance of the model at the end of the first stage of training.}}
\vspace{.5em}
\resizebox{1.\linewidth}{!}
{
\begin{tabular}{lcccccc}
\toprule[1.25pt]
 {\textbf{Methods}} & \textbf{\#Activated Heads (\%)} & {\textbf{MMLU (5)}} & {\textbf{CEVAL (5)}} & {\textbf{CMMLU (5)}} & {\textbf{GSM8K(8)}} & {\textbf{TruthfulQA}} \\ 
 \midrule
 LLaMA3-8B~\tiny{\citep{dubey2024llama}}  & 100 & 65.2 & 52.3 & 50.7 & 49.5 & 35.4 \\
 \cmidrule(rl){1-2} \cmidrule(rl){3-7}
 \rowcolor{aliceblue!60} \textbf{MoH-LLaMA3-8B} & 75 & 65.8 & 61.5 & 64.4 & 56.9 & 44.0 \\
 \midrule
 \midrule
 {\textbf{Methods}} & \textbf{\#Activated Heads (\%)} & {\textbf{HellaSwag (10)}} & {\textbf{LogiQA}} & {\textbf{BoolQ (32)}} & {\textbf{LAMBADA}} & {\textbf{SciQ}} \\ 
 \midrule
 LLaMA3-8B~\tiny{\citep{dubey2024llama}}  & 100 & 81.9 & 30.0 & 83.9 & 75.5 & 94.0 \\
 \cmidrule(rl){1-2} \cmidrule(rl){3-7}
 \rowcolor{aliceblue!60} \textbf{MoH-LLaMA3-8B} & 75 & 80.1 & 30.3 & 84.0 & 76.4 & 92.2 \\
 \midrule
 \midrule
 {\textbf{Methods}} & \textbf{\#Activated Heads (\%)} & {\textbf{PIQA}} & {\textbf{WinoGrande}} & {\textbf{NQ (32)}} & {\textbf{ARC-C (25)}} & {\textbf{Average}} \\ 
 \midrule
 LLaMA3-8B~\tiny{\citep{dubey2024llama}}  & 100 & 81.0 & 72.5 & 31.5 & 59.0 & 61.6 \\
 \cmidrule(rl){1-2} \cmidrule(rl){3-7}
 \rowcolor{aliceblue!60} \textbf{MoH-LLaMA3-8B} & 75 & 78.8 & 72.9 & 28.3 & 60.1 & \textbf{64.0} \\
\bottomrule[1.25pt]
\end{tabular}
}
\vspace{-.5em}
\label{tab: llama3}
\end{table*}

\myparagraph{Results.} As shown in Tab.~\ref{tab: vit}, despite activating only a subset of attention heads, MoH-ViT achieves highly competitive performance compared to current state-of-the-art methods. For example, MoH-ViT-B achieves 84.9\% Top-1 accuracy on the ImageNet-1K classification benchmark with just 75\% of the attention head. In contrast, the well-established ViT baseline, TransNeXt, attains a slightly lower accuracy of 84.8\% while requiring 100\% of the heads to be activated. These results suggest that MoH is a promising alternative to multi-head attention for vision model design.

\begin{figure*}[tp]
\centering
\includegraphics[width=1\textwidth]{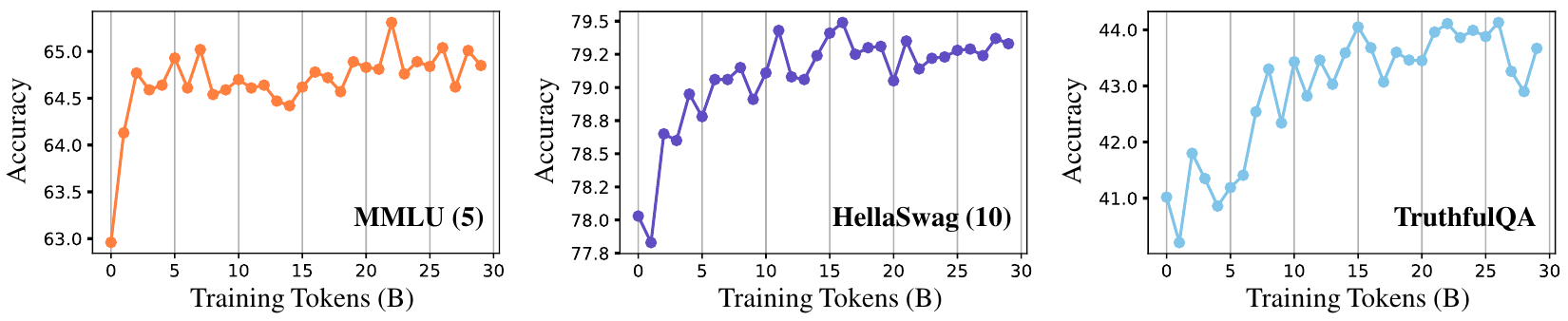}
\vspace{-2.em}
\caption{\textbf{Performance evolution during continue-tuning.} The MoH model quickly recovers to over 95\% of the performance of the original model within a training budget of 10B tokens. Then, the performance gradually improves with the increase of the training tokens.}
\vspace{-.5em}
\label{fig2}
\end{figure*}

\begin{table*}[t]
\footnotesize
\centering
\caption{\textbf{Ablation study on the impact of each component of the proposed MoH.} The image classification results are from MoH-ViT-S, by utilizing 75\% of the attention heads with a training budget of 100 epochs. The class-conditional image generation results come from MoH-DiT-S/2-400K, also by using 75\% of the attention heads, with a training budget of 400K training steps.}
\vspace{.5em}
\setlength{\tabcolsep}{12.9pt}
{
\begin{tabular}{ccccccccccc}
\toprule[1.25pt]
 {\textbf{Shared}} & {\textbf{Two-Stage}} & {\textbf{Image Classification}} &\multicolumn{5}{c}{\textbf{Class-Conditional Image Generation}} \\ \cmidrule(rl){3-3} \cmidrule(rl){4-8}
 \textbf{Heads} & \textbf{Routing} & \textbf{Acc (\%)$\uparrow$} & {\textbf{FID$\downarrow$}} & {\textbf{sFID$\downarrow$}} & {\textbf{IS$\uparrow$}} & {\textbf{Precision$\uparrow$}} & {\textbf{Recall$\uparrow$}} \\ 
 \midrule
 & & 75.6 & 71.97 & 13.58 & 19.06 & 0.35 & \textbf{0.55} \\ 
 \checkmark & & 78.3 & 69.54 & \textbf{12.80} & 19.67 & \textbf{0.36} & \textbf{0.55} \\
 \rowcolor{aliceblue!60} \checkmark & \checkmark & \textbf{78.6} & \textbf{69.42} & 12.85 & \textbf{19.96} & \textbf{0.36} & \textbf{0.55} \\
\bottomrule[1.25pt]
\end{tabular}
}
\vspace{-.5em}
\label{tab: ablation0}
\end{table*}

\begin{table*}[t]
\footnotesize
\centering
\caption{\textbf{Ablation study on the impact of the shared heads ratio among activated heads.} All results are from MoH-ViT-S, by using 75\% of the heads with a training budget of 100 epochs.}
\vspace{.5em}
\setlength{\tabcolsep}{9.4pt}
{
\begin{tabular}{lccccccccc}
\toprule[1.25pt]
 {\textbf{Ratio of Shared Heads}} & 13.9\% & 27.6\% & 31.3\% & 35.9\% & 37.5\% & 40.5\% & 46.8\% & 60.4\% & 74.0\% \\ 
 \midrule
 {\textbf{Accuracy (\%)}} & 78.6 & 78.5 & 78.4 & 78.4 & 78.5 & 78.6 & 78.4 & 78.6 & 78.4 \\
\bottomrule[1.25pt]
\end{tabular}
}
\label{tab: ablation1}
\vspace{-0.5em}

\centering
\caption{\textbf{Comparisons about inference time.} We convert the $\bm{Q}$, $\bm{K}$, and $\bm{V}$ features into sparse matrices using the mask generated by the router and replace the dense matrix multiplication in the attention mechanism with sparse matrix multiplication. To eliminate the impact of underlying operator optimizations, we replaced all matrix multiplications with sparse matrix multiplication when testing for speed.}
\vspace{.5em}
{
\begin{tabular}{lccccccccc}
\toprule[1.25pt]
 {\textbf{Methods}} & \textbf{\#Head Num} & \textbf{\#Head Dim} & \textbf{\#Sequence Length}	 &  \textbf{\#Activated Heads (\%)} & \textbf{Time (ms)} \\ 
 \midrule
 Multi-Head Attention & 32 & 64 & 256 & 100 & 0.360 \\
 \textbf{MoH (Ours)} & 32 & 64 & 256 & 90 & 0.352 \\
 \textbf{MoH (Ours)} & 32 & 64 & 256 & 75 & 0.321 \\
 \rowcolor{aliceblue!60} \textbf{MoH (Ours)} & 32 & 64 & 256 & 50 & \textbf{0.225} \\
 \midrule
 \midrule
 Multi-Head Attention & 32 & 64 & 512 & 100 & 1.376 \\
 \textbf{MoH (Ours)} & 32 & 64 & 512 & 90 & 1.351 \\
 \textbf{MoH (Ours)} & 32 & 64 & 512 & 75 & 1.180 \\
 \rowcolor{aliceblue!60} \textbf{MoH (Ours)} & 32 & 64 & 512 & 50 & \textbf{0.863} \\
\bottomrule[1.25pt]
\end{tabular}
}
\vspace{-.5em}
\label{tab: ablation2}
\end{table*}

\subsection{DiT for Class-Conditional Image Generation}
\myparagraph{Model Settings.} For Diffusion models with Transformers~(DiT)~\citep{peebles2023scalable}, we only replace the standard multi-head attention with our MoH in MoH-DiT models, while keeping all other training parameters identical to DiT. We use the ImageNet-1K dataset for class-conditional image generation at a resolution of 256$\times$256. To evaluate generation performance, we use Frechet Inception Distance~(\textbf{FID})~\citep{heusel2017gans} to assess overall sample quality, \textbf{Precision} and \textbf{Recall}~\citep{kynkaanniemi2019improved} to measure fidelity and diversity separately, and \textbf{sFID}~\citep{nash2021generating} as a metric that better captures spatial relationships than FID. Moreover, we use Inception Score (\textbf{IS})~\citep{salimans2016improved} as another metric for fidelity.

\myparagraph{Training Details.} Following DiT, the final linear layer is initialized with zeros, and all other layers follow standard ViT weight initialization. We train all models using the AdamW optimizer~\citep{loshchilov2017decoupled} with a constant learning rate of 1e-4, no weight decay, and a batch size of 256, applying horizontal flips for data augmentation. Following DiT, we employ the Exponential Moving Average (EMA) of MoH-DiT weights during training with a decay rate of 0.9999, generating all images using the EMA model. We use an off-the-shelf pre-trained variational autoencoder~\citep{kingma2013auto} model from Stable Diffusion~\citep{rombach2022high}. Following TransNeXt, our attention-head activation budget is unevenly distributed across layers, with fewer attention heads activated in the shallow layers and more in the deeper layers.

\myparagraph{Results.} As shown in Tab.~\ref{tab: dit2}, MoH-DiT models consistently outperform DiT models with 90\% of heads activated. However, when only 75\% of the heads are activated, MoH-DiT models perform worse than DiT models. This may be because image generation tasks are dense prediction tasks that require attention mechanisms to capture pixel-level fine-grained relationships, leaving less redundancy in the attention heads compared to image classification tasks. These results suggest that MoH is a promising alternative to multi-head attention for diffusion models.

\subsection{Training LLMs from Scratch}
\myparagraph{Model Settings.} For training LLMs from scratch, we use Megatron~\citep{shoeybi2019megatron}, an open-source training code, as the training framework. Please refer to the Appendix for detailed hyper-parameter settings~(Tab.~\ref{tab: LLMs settings}) of various MoH-LLMs. The evaluation is performed on multiple benchmarks using the Eleuther AI Language Model Evaluation Harness~\citep{eval-harness}, a unified framework for testing generative language models. Since the parameters are only about 0.2B for the smallest model, we select 6 simple benchmarks as the metric. Specifically, we report 0-shot accuracy on \textbf{SciQ}~\citep{welbl2017crowdsourcing}, \textbf{PIQA}~\citep{bisk2020piqa}, \textbf{WinoGrande}~\citep{sakaguchi2021winogrande}, \textbf{OpenbookQA}~\citep{OpenBookQA2018}, \textbf{LogiQA}~\citep{liu2020logiqa}, and \textbf{TruthfulQA}~\citep{lin-etal-2022-truthfulqa}.

\myparagraph{Training Details.} We only use public datasets for training, ensuring accessibility for academic research. Specifically, we sample from the \textbf{RedPajama}~\citep{together2023redpajama}, \textbf{Dolma}~\citep{dolma}, and \textbf{Pile}~\citep{pile} datasets according to different sampling probabilities. Please refer to the Appendix for detailed sample ratios. Following previous works~\citep{jin2024moe++}, we utilize the tokenizer from LLaMA2~\citep{touvron2023llama2}, which contains 65,536 vocabulary tokens.

\myparagraph{Results.} As shown in Tab.~\ref{tab: LLMs}, despite activating only a subset of attention heads, MoH-LLMs achieve highly competitive performance compared to our baseline models. For example, MoH-LLM-S achieves an average accuracy of 45.4\% with just 50\% of the attention heads activated. In contrast, the baseline model reaches a slightly lower accuracy of 43.9\% with 100\% of the attention heads activated. These results suggest that MoH is a promising alternative to vanilla multi-head attention for training LLMs from scratch. Surprisingly, we find that for MoH-LLM-S, activating only 50\% of the attention heads outperforms activating 75\%. We consider it may be because when both the model and dataset are small, activating fewer heads effectively regularizes the model. However, as the amount of data increases, activating more heads offers a higher potential for performance.

\subsection{Continue-Tuning LLaMA3-8B}
\myparagraph{Model Settings.} To significantly enhance the applicability of the proposed MoH method, we also attempt to further continue-tune pre-trained multi-head attention models, such as LLaMA3-8B, into MoH models. However, this presents three challenges. \textbf{(i) Determining the shared attention heads}: We simply select the first 16 attention heads of each layer as shared heads. \textbf{(ii) Adding head routers}: Integrating a randomly initialized router into the pre-trained model without compromising its original performance requires careful training techniques. To address this, we propose a parameter-free router that determines routing scores using the $\ell_2$ norm of the query of each attention head. \textbf{(iii)~Weighting attention heads}: We observe that weighting the attention head outputs significantly alters the distribution of the output of the attention layer, which necessitates a large amount of training data to restore the original performance. To tackle this, we quantize the routing score and use the straight-through estimator~\citep{bengio2013estimating,liu2022nonuniform} to back-propagate the gradients through the sparsity function. Specifically, given the input token $\bm{x}$, we employ a quantizer for activation routing scores, with its forward pass formulated as:
\begin{equation}
g_{i}^{q} = \mathds{1}(\text{Token $\bm{x}$ selects Head $i$}),
\end{equation}
where $\mathds{1}(*)$ denotes the indicator function. $g_{i}^{q}$ represents the quantized routing score. We then adopt a straight-through estimator, which assigns the incoming gradients to a threshold operation to be the outgoing gradients:
\begin{equation}
\frac{\partial\mathcal{L}}{\partial g_{i}^{q}} = \frac{\partial\mathcal{L}}{\partial g_{i}},
\end{equation}
where $g_{i}$ denotes the real-valued routing score. This approximation function significantly mitigates the issue of gradient vanishing~\citep{wang2024q}. Similar to training LLMs from scratch, we also use Megatron~\citep{shoeybi2019megatron}, an open-source training code, as the training framework.

\begin{figure*}[tp]
\centering
\includegraphics[width=1\textwidth]{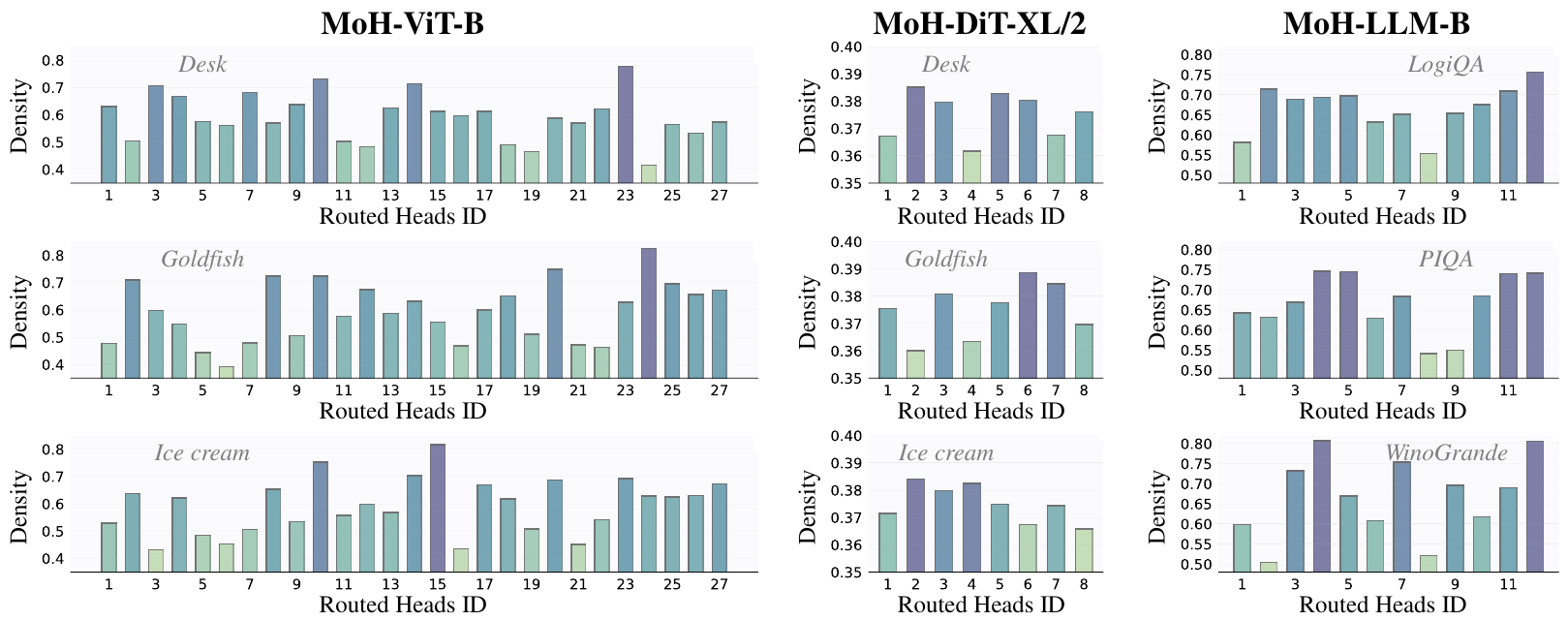}
\vspace{-2.em}
\caption{\textbf{Visualization of the head load distribution in the final MoH layer.} For ViT and DiT, we present the head load distributions for the categories ``Desk'', ``Goldfish'', and ``Ice cream''. For LLM, we display the head distributions for the tasks ``LogiQA'', ``PIQA'', and ``WinoGrande''. MoH-ViT-B, MoH-DiT-XL/2, and MoH-LLM-B activate 75\%, 90\%, and 75\% of the attention heads, respectively. {``Density'' denotes the ratio of the number of head activations to the total number of tokens.}}
\vspace{-.5em}
\label{fig3}
\end{figure*}

\myparagraph{Training Details.} We find that if there is a discrepancy between the continue-training data and the original training data distribution of the model, the performance of the model may fluctuate wildly at the beginning of the training process. Since we are unable to have access to the raw training data of LLaMA3, we address these potential performance fluctuations by dividing the training process into two stages. In the first stage, we continue-tune the original LLaMA3-8B model using 300B tokens to adapt the model to our dataset. In the second stage, we continue-tune this adapted model into our proposed MoH model with 100B tokens. We utilize the lm-evaluation-harness package to assess performance on a comprehensive suite of downstream tasks: (i) Following Pythia~\citep{biderman2023pythia}, we report 0-shot accuracy on \textbf{LAMBADA}~\citep{paperno2016lambada}, \textbf{LogiQA}~\citep{liu2020logiqa}, \textbf{PIQA}~\citep{bisk2020piqa}, \textbf{SciQ}~\citep{welbl2017crowdsourcing}, and \textbf{WinoGrande}~\citep{sakaguchi2021winogrande}. (ii) We report the accuracy of Chinese tasks, including 5-shot \textbf{CEVAL}~\citep{huang2024c} and 5-shot \textbf{CMMLU}~\citep{li2023cmmlu}. (iii) We report the accuracy of tasks from the Open LLM Leaderboard~\citep{open-llm-leaderboard-v1}, including 10-shot \textbf{HellaSwag}~\citep{zellers2019hellaswag}, 25-shot \textbf{ARC Challenge (ARC-C)}~\citep{clark2018think}, and 5-shot \textbf{MMLU}~\citep{hendrycks2020measuring}. (iv) We report the exact match score for 32-shot \textbf{Natural Questions (NQ)}~\citep{kwiatkowski2019natural} and the accuracy for 32-shot \textbf{BoolQ}~\citep{clark2019boolq}. (v) We report the exact match score for 8-shot \textbf{GSM8K}~\citep{cobbe2021training} to evaluate the math ability. (vi) Moreover, we report 0-shot accuracy on \textbf{TruthfulQA}~\citep{lin-etal-2022-truthfulqa} to assess the ability to generate truthful answers.

\myparagraph{Results.} As shown in Fig.~\ref{fig2}, MoH-LLaMA3-8B quickly recovers to over 95\% of the performance of the original model within a training budget of 10B tokens. After continue-tuning with 100B tokens, as shown in Tab.~\ref{tab: llama3}, MoH-LLaMA3-8B achieves an average accuracy of 64.0\% across 14 benchmarks, outperforming LLaMA3-8B by 2.4\% by utilizing only 75\% of the attention heads. These results demonstrate that pre-trained multi-head attention models can be further continue-tuned into our MoH models, significantly enhancing the applicability of the MoH method.

\subsection{Ablative Analysis}
\myparagraph{Effect of Each Component of the Proposed MoH.} 
To explore the impact of each component of our MoH method, we provide the ablation results in Tab.~\ref{tab: ablation0}. ``Shared Heads'' refers to a subset of attention heads that are always activated. ``Two-Stage Routing'' represents the dynamic coefficient that balances the weights between shared and routed heads over the routing score, as described in Eq.~\ref{router} and Eq.~\ref{alpha}. As shown in Tab.~\ref{tab: ablation0}, shared heads significantly improve model performance by effectively capturing common knowledge, allowing the routed heads to focus more on domain-specific information. Moreover, two-stage routing further enhances model performance by dynamically balancing the weights between shared and routed heads. Our full model achieves the best performance, demonstrating that both components significantly benefit the attention mechanism.

\myparagraph{Effect of the Shared Heads Ratio among Activated Heads.} 
In Tab.~\ref{tab: ablation1}, we provide the ablation study on the shared heads ratio among activated heads. We find that model performance remains relatively consistent across a wide range of shared heads ratios (from 13.9\% to 74.0\%). These results indicate that the performance of the model is stable as long as the shared heads ratio is not extreme. From another perspective, shared heads can be viewed as a form of Soft MoE~\citep{puigcerver2023sparse}. Based on the findings from the Soft MoE paper~\citep{puigcerver2023sparse}, we recommend using a higher ratio of shared heads among the activated heads (greater than 40\%).

\section{Discussion}
\myparagraph{The Efficiency of Our Proposed MoH.}
To explore if our method performs better with longer sequences, we increase the input sequence length. For rows 1 to 4 of Tab.~\ref{tab: ablation2}, the input length is 256. For rows 5 to 8, it is 512. As shown in Tab.~\ref{tab: ablation2}, although dynamic routing introduces additional computational overhead, MoH still outperforms standard multi-head attention mechanisms. Furthermore, as the input sequence gets longer, the advantage of MoH grows.

\myparagraph{Visualization of the Head Load Distribution.}
As shown in Fig.~\ref{fig3}, we observe significant variation in attention head assignments across different categories and task topics, indicating that the MoH model adapts to diverse tasks by employing distinct head assignment patterns. This characteristic of MoH allows different attention heads to focus on different types of tasks, making parameter utilization more efficient than multi-head attention. For additional visualizations of MoH-LLaMA3-8B and a detailed analysis of the head load distribution, please refer to Appendix~\ref{appendix:Additional Qualitative Analysis}.

\myparagraph{The Difference between MoH and MoA.} We clarify the differences between MoH and MoA~\citep{zhang2022mixture} from the following three aspects. \textbf{First, in terms of motivation}, the goal of MoH is to improve the efficiency and performance of the attention mechanism without increasing the number of parameters. In contrast, MoA shares the motivation of MoE, which is to expand model parameters while keeping inference costs low. Therefore, the model settings of MoH are more stringent than those of MoA. \textbf{Second, in terms of methodology}, our MoH introduces shared heads and two-stage routing to enhance the standard MoE method. More importantly, we show that pre-trained multi-head attention models can be further continue-tuned into our MoH models, greatly improving the applicability of the proposed MoH method. In contrast, MoA directly combines multi-head attention with MoE. Due to the adoption of shared keys and values, MoA must be trained from scratch, which limits its applicability. \textbf{Finally, in terms of model frameworks}, our MoH is validated across various popular model frameworks and tasks, including ViT, DiT, and decoder-only LLMs, while MoA is only validated for language tasks.

\section{Conclusion}
In this work, we introduce MoH, a promising alternative to multi-head attention. MoH enables each token to adaptively select the appropriate attention heads, improving both model performance and inference efficiency without increasing the number of parameters. Extensive experiments across various popular model frameworks, including ViT, DiT, and LLMs, demonstrate that MoH outperforms multi-head attention, even when using only 50\%$\sim$90\% of the attention heads. This work represents a promising step toward advanced and efficient attention-based models, which may be helpful to both the research and industrial communities.

\section*{Acknowledgements}
This work was supported in part by the Natural Science Foundation of China (No. 62202014, 62332002, 62425101, 62088102), and NUS Start-up Grant A-0010106-00-00. Besides, this work was performed when Peng Jin was an Intern at Skywork AI.

\section*{Impact Statement}
This work is an important step toward creating more advanced and efficient attention-based models, which could benefit both the research and industrial communities. Efficient attention models will not only lower the training costs for researchers but also greatly reduce the expenses involved in deploying and using large models.

\bibliography{main}
\bibliographystyle{icml2025}

\newpage
\appendix
\onecolumn
\renewcommand{\thefootnote}{\fnsymbol{footnote}}
\renewcommand{\thetable}{\Alph{table}}
\renewcommand{\theequation}{\Alph{equation}}
\renewcommand{\thefigure}{\Alph{figure}}

\setcounter{table}{0}
\setcounter{section}{0}
\setcounter{figure}{0}
\setcounter{equation}{0}

\myparagraph{Abstract.} This appendix provides additional discussions~(Appendix~\ref{appendix:Additional Discussions}), implementation details~(Appendix~\ref{Experimental Setup1}), several additional experiments~(Appendix~\ref{appendix:Additional Experiments}), more qualitative analysis~(Appendix~\ref{appendix:Additional Qualitative Analysis}), and details of quantitative evaluations for LLMs~(Appendix~\ref{appendix:Details of Quantitative Evaluations}).

\section{Additional Discussions}\label{appendix:Additional Discussions}
\subsection{Why is MoH Superior to Vanilla Multi-Head Attention?}
We demonstrate that MoH is superior to vanilla multi-head attention from both theoretical and experimental perspectives.

Specifically, MoH not only improves efficiency and model performance but also helps different attention heads to specialize better compared to multi-head attention.

From the theoretical perspective, in standard multi-head attention, 
all heads use the same data, which can cause them to learn similar features. 
Many studies have pointed out that there are redundant heads in multi-head attention.
Given a minibatch of data $D$, the gradient of each attention head in multi-head attention can be written as $\mathbb{E}_{x \in D} [\frac{\partial \mathcal{L}(x)}{\partial h_{i} }]$.

In contrast, in MoH, routed heads are trained only on smaller subsets of data specifically assigned to them. 
In MoH's routing mechanism, the data is divided into $h-h_s$ subsets $\{D_1,D_2,...,D_{h-h_s}\}$, with each subset corresponding to a routed head. 
Besides, the routing score for each attention head acts as an adaptive adjustment to the learning rate, enabling the attention heads in MoH to specialize more effectively.
Given a minibatch of data $D$ and the router $G(*)$, the gradient of each routed head in MoH can be written as $ \mathbb{E}_{x \in D_i } [G(x)_i \frac{\partial \mathcal{L}(x)}{\partial h_{i} }] $.
The gradient of each shared head in MoH can be written as $ \mathbb{E}_{x \in D } [G(x)_i \frac{\partial \mathcal{L}(x)}{\partial h_{i} }] $. As shown in Tab.~\ref{apxtab: advantage 1}, the routing mechanism and adaptive weights in MoH enable attention heads to specialize more effectively compared to standard multi-head attention.

\begin{table*}[ht]
\footnotesize
\centering
\caption{\textbf{Comparisons between the multi-head attention and our proposed mixture-of-head attention.}}
\vspace{.5em}
{
\begin{tabular}{lcccc}
\toprule[1.25pt]
 {\textbf{Methods}} & \textbf{\#Head Type} & {\textbf{\#Data}} & {\textbf{\#Weight (learning rate)}} & {\textbf{\#Gradient}}  \\ 
 \midrule
 Multi-Head Attention & - & $D$ & 1 & $ \mathbb{E}_{x \in D} [\frac{\partial \mathcal{L}(x)}{\partial h_{i} }]$  \\
 MoH & routed head & $D_i \in D$ & $G(x)_i$ & $ \mathbb{E}_{x \in D_i } [G(x)_i \frac{\partial \mathcal{L}(x)}{\partial h_{i} }] $  \\
 MoH & shared head & $D$ & $G(x)_i$ & $ \mathbb{E}_{x \in D } [G(x)_i \frac{\partial \mathcal{L}(x)}{\partial h_{i} }]$  \\
\bottomrule[1.25pt]
\end{tabular}
}
\label{apxtab: advantage 1}
\end{table*}

From the experimental perspective, we calculated the similarity of attention patterns and output features of different attention heads (include routed heads and shared heads). As shown in Tab.~\ref{apxtab: advantage 2}, the similarity of attention patterns and output features among attention heads in MoH is lower than in standard multi-head attention, indicating reduced redundancy and greater differentiation among the attention heads in MoH.

\begin{table*}[ht]
\footnotesize
\centering
\caption{\textbf{The similarity of attention patterns and output features among attention heads.} Given a pair of attention score matrices $A$ and $A'$, we calculate the similarity of attention patterns as $ 1 - \frac{1}{2} \mathbb{E}[ ||A-A'||_{1} ] $. Since attention scores form a probability distribution for each query, the similarity is always between 0 to 1.}
\vspace{.5em}
{
\begin{tabular}{lcccc}
\toprule[1.25pt]
 \multirow{2}{*}{\textbf{Methods}} &\multicolumn{2}{c}{\textbf{Similarity of Attention Patterns}}  &\multicolumn{2}{c}{\textbf{Cosine Similarity of Output Features}}   \\ \cmidrule(rl){2-3} \cmidrule(rl){4-5}
 &  \textbf{ViT} &  \textbf{LLM} &  \textbf{ViT} &  \textbf{LLM}  \\ 
 \midrule
 Multi-Head Attention & 0.5159 & 0.4795 & 0.0411 & 0.2550  \\
 \textbf{MoH} & \textbf{0.3978} & \textbf{0.4333} & \textbf{0.0165} & \textbf{0.2042}  \\
\bottomrule[1.25pt]
\end{tabular}
}
\label{apxtab: advantage 2}
\end{table*}

\subsection{Limitations and Future Work}
In this section, we delineate the limitations of our work and outline avenues for future research.

\myparagraph{Heterogeneous Attention Heads.}
We find that different attention heads operate in parallel within the attention mechanism, suggesting that different heads can have varying hidden sizes. Future work could explore the use of heterogeneous attention heads based on our MoH framework.

\myparagraph{Lower Activation Rate.}
Currently, MoH outperforms multi-head attention by utilizing only 50\%$\sim$90\% of the attention heads. However, this is still a relatively high proportion. Future work could aim to further optimize MoH, reducing head activation to less than 50\%.

\myparagraph{Multimodal Inputs.}
Effectively processing information from multiple modalities in the attention mechanism remains an open question. Recent work~\citep{wan2024look} has shown that visual and textual tokens exhibit distinct attention patterns in multi-head attention. Future work could explore the attention patterns of MoH with different modal inputs, for example within multimodal large language models~\citep{jin2024chat,lin2023video,lin2024moe,liusphinx,jin2023video,jin2024hierarchical}.

\myparagraph{More Downstream Tasks.}
We evaluate our proposed MoH across various popular model frameworks, including ViT for image classification, DiT for class-conditional image generation, and LLMs for language tasks. Future work can explore the application of MoH in more downstream tasks, such as audio tasks and multimodal tasks.

\myparagraph{More Parameters.}
Due to computational constraints, the maximum number of MoH model parameters in our experiments is limited to 8B (MoH-LLaMA3-8B). However, our MoH method is highly generalizable and can be scaled to larger models in future research.

\section{Implementation Details}\label{Experimental Setup1}
\subsection{ViT for Image Classification}
\myparagraph{Training Details.} Our MoH-ViT models are trained for 300 epochs using automatic mixed precision across 8 GPUs. We follow the training strategy of TransNeXt, which includes various data augmentation techniques, including Random Augmentation~\citep{cubuk2020randaugment}, Mixup~\citep{zhang2017mixup}, CutMix~\citep{yun2019cutmix}, and Random Erasing~\citep{zhong2020random}. We also apply Label Smoothing~\citep{szegedy2016rethinking} and DropPath~\citep{huang2016deep} to regularize our models. We optimize our models using AdamW optimizer~\citep{loshchilov2017decoupled} with a gradient clipping norm of 1.0 and a weight decay of 0.05. The initial learning rate is set to 1e-3, with a 5-epoch warm-up starting at 1e-6. A cosine learning rate scheduler~\citep{loshchilov2016sgdr} is employed to decay the learning rate. During training, images are randomly cropped to a size of 224$\times$224. It is worth noting that we do not use Exponential Moving Average~(EMA) weights.

\subsection{DiT for Class-Conditional Image Generation}
\myparagraph{Training Details.} Following DiT, the final linear layer is initialized with zeros, and all other layers follow standard ViT weight initialization. We train all models using the AdamW optimizer~\citep{loshchilov2017decoupled} with a constant learning rate of 1e-4, no weight decay, and a batch size of 256, applying horizontal flips for data augmentation. Following DiT, we employ the Exponential Moving Average (EMA) of MoH-DiT weights during training with a decay rate of 0.9999, generating all images using the EMA model. We use an off-the-shelf pre-trained variational autoencoder~\citep{kingma2013auto} model from Stable Diffusion~\citep{rombach2022high}. Following TransNeXt, our attention-head activation budget is unevenly distributed across layers, with fewer attention heads activated in the shallow layers and more in the deeper layers.

\subsection{Training LLMs from Scratch}
\myparagraph{Model Settings.} For training LLMs from scratch, we use Megatron~\citep{shoeybi2019megatron}, an open-source training code, as the training framework. The detailed hyper-parameter settings of various MoH-LLMs are shown in Tab.~\ref{tab: LLMs settings}.

\begin{table*}[ht]
\footnotesize
\centering
\caption{\textbf{Sizes and architectures of MoH-LLMs and LLMs.} ``MoH-LLM-B'' has more parameters than ``LLM-B'' due to the additional parameters introduced by the router network.}
\vspace{.5em}
{
\begin{tabular}{lccccccccc}
\toprule[1.25pt]
 {\textbf{Methods}} & \textbf{\#Params} & {\textbf{\#Layers}} & {\textbf{\#Hidden Size}} & {\textbf{\#Intermediate Size}} & {\textbf{\#Heads}} & {\textbf{\#Head Dim}}   \\ 
 \midrule
 LLM-S & 186 & \multirow{2}{*}{12} & \multirow{2}{*}{768} & \multirow{2}{*}{2048} & \multirow{2}{*}{12} & \multirow{2}{*}{64} & \\
 MoH-LLM-S & 186 & & & & & \\
 \midrule
 LLM-B & 881 & \multirow{2}{*}{24} & \multirow{2}{*}{1536} & \multirow{2}{*}{4096} & \multirow{2}{*}{16} & \multirow{2}{*}{96} \\
 MoH-LLM-B & 882 & & & & & \\
\bottomrule[1.25pt]
\end{tabular}
}
\label{tab: LLMs settings}
\end{table*}

\myparagraph{Data Details.} Consistent with previous works, we use the tokenizer of LLaMA2, which contains 65,536 vocabulary tokens. It is worth noting that MoH-LLM is trained exclusively on public datasets, making it accessible for academic research settings. Tab.~\ref{apdx tab: datasets} shows the detailed sample ratios of different open-source datasets. Specifically, we sample from the following datasets according to different sampling probabilities: 

\begin{itemize}
    \item The \textbf{RedPajama}~\citep{together2023redpajama} includes training data from seven domains: CommonCrawl, C4, Github, Wikipedia, Books, ArXiv, and StackExchange.

    \item The \textbf{Dolma}~\citep{dolma}, a large and diverse open English text corpus, contains 3 trillion tokens sampled from seven sources, including web pages from Common Crawl, code from The Stack, curated web data from C4~\citep{raffel2020exploring}, social media conversations from Reddit, academic papers from PeS2o, public domain books from Project Gutenberg, and comprehensive content from Wikipedia and Wikibooks.

    \item The \textbf{Pile}~\citep{pile}, an open-source English text corpus for training large language models, includes 22 diverse, publicly available datasets such as Wikipedia, NIH ExPorter, ArXiv, Books3, BookCorpus2, OpenSubtitles, YoutubeSubtitles, and Enron Emails.
\end{itemize}

\begin{table*}[ht]
\footnotesize
\centering
\caption{\textbf{Sampling ratio of different open-source datasets for MoH-LLMs.} MoH-LLM is trained exclusively on public datasets, making it accessible for academic research settings.}
\vspace{.5em}
{
\begin{tabular}{lc}
\toprule[1.25pt]
  & {\textbf{Sampling Ratio}} \\ 
 \midrule
 Redpajama Books & 4.24\%  \\
 Redpajama Wikipedia & 3.50\%  \\
 Redpajama ArXiv & 4.37\%  \\
 Redpajama StackExchange & 3.19\%  \\
 Redpajama C4 & 10.94\%  \\
 \midrule
 Dolma & 61.28\% \\
 \midrule
 Pile & 12.48\% \\
\bottomrule[1.25pt]
\end{tabular}
}
\label{apdx tab: datasets}
\end{table*}

\myparagraph{Training Hyper-Parameters.} 
Tab.~\ref{apdx tab: hyper-parameters} shows the detailed training hyper-parameters of MoH-LLMs. Specifically, all MoH-LLMs are trained with the AdamW optimizer~\citep{loshchilov2017decoupled}, using a batch size of 4 million tokens with a sequence length of 2048. The final learning rate is set to 10\% of the maximum. During training, a weight decay of 0.1 and gradient clipping of 1.0 are applied. For LLM-S and MoH-LLM-S, the maximum learning rate is set to 3e-4. For LLM-B and MoH-LLM-B, the maximum learning rate is set to 5e-4.

\begin{table*}[ht]
\footnotesize
\centering
\caption{\textbf{Training hyper-parameters of MoH-LLMs.}}
\vspace{.5em}
{
\begin{tabular}{lccc}
\toprule[1.25pt]
  & {\textbf{MoH-LLM-S} \type{100B}} & {\textbf{MoH-LLM-B} \type{100B}} & {\textbf{MoH-LLM-B} \type{200B}}\\ 
  & {(\textbf{LLM-S} {\type{100B}})} & {(\textbf{LLM-B} {\type{100B}})} & {(\textbf{LLM-B} {\type{200B}})}\\ 
 \midrule
 Training budget & 100B & 100B & 200B \\
 Maximum learning rate & 3e-4 & 5e-4 & 5e-4 \\
 Final learning rate & 3e-5 & 5e-5 & 5e-5 \\
 LR warmup init & 1e-7 & 1e-7 & 1e-7 \\
 LR warmup iters & 2000 & 500 & 500 \\
 Sequence length & 2048 & 2048 & 2048 \\
 Batch size (tokens) & 4M & 4M & 4M \\
 $\beta$ for $\mathcal{L}_{b}$ & 0.01 & 0.01 & 0.01 \\
 Tensor parallel & 1 & 1 & 1 \\
 Pipeline parallel & 1 & 1 & 1 \\
\bottomrule[1.25pt]
\end{tabular}
}
\label{apdx tab: hyper-parameters}
\end{table*}

\subsection{Continue-Tuning LLaMA3-8B}
\myparagraph{Training Hyper-Parameters.} 
Tab.~\ref{apdx tab: hyper-parameters 1} shows the detailed training hyper-parameters of MoH-LLaMA3-8B. We find that if there is a discrepancy between the continue-training data and the original training data distribution of the model, the performance of the model may fluctuate wildly at the beginning of the training process. Since we do not have access to the raw training data of LLaMA3, we address these potential performance fluctuations by dividing the training process into two stages. In the first stage, we continue-tune the original LLaMA3-8B model using 300B tokens to adapt it to our dataset. In addition, during the first stage, to enhance the Chinese ability of the model, we expand the vocabulary size. Specifically, we increase the original LLaMA3-8B vocabulary size from 128,256 to 160,896. In the second stage, we continue-tune this adapted model into our proposed MoH model with 100B tokens. During the first stage, the maximum learning rate is set to 6e-5, and the final learning rate is 6e-6. In the second stage, the maximum learning rate is set to 2e-5, and the final learning rate is 1e-6. For both stages, we employ the AdamW optimizer~\citep{loshchilov2017decoupled}, with a batch size of 16 million tokens with a sequence length of 8192. During training, we use a weight decay of 0.1 and gradient clipping of 1.0.

\begin{table*}[ht]
\footnotesize
\centering
\caption{\textbf{Training hyper-parameters of MoH-LLaMA3-8B.} We divide the training process into two stages. In the first stage, we continue-tune the LLaMA3-8B model using 300B tokens. In the second stage, we continue-tune this adapted model into our proposed MoH model with 100B tokens.}
\vspace{.5em}
{
\begin{tabular}{lccc}
\toprule[1.25pt]
  & {\textbf{The First Stage}} & {\textbf{The Second Stage}} \\ 
 \midrule
 Training budget & 300B & 100B \\
 Maximum learning rate & 6e-5 & 2e-5 \\
 Final learning rate & 6e-6 & 1e-6 \\
 LR warmup iters & 50 & 50 \\
 Sequence length & 8192 & 8192 \\
 Batch size (tokens) & 16M & 16M \\
 $\beta$ for $\mathcal{L}_{b}$ & - & 0.01 \\
 Tensor parallel & 2 & 1 \\
 Pipeline parallel & 1 & 8 \\
\bottomrule[1.25pt]
\end{tabular}
}
\label{apdx tab: hyper-parameters 1}
\end{table*}

\begin{table*}[ht]
\footnotesize
\centering
\caption{\textbf{{Comparisons between MoH-LLaMA3-8B and LLaMA3-8B-stage1.}} {MoH-LLaMA3-8B outperforms LLaMA3-8B-stage1 by utilizing only 75\% of the attention heads.}}
\vspace{.5em}
{
\begin{tabular}{lcccccc}
\toprule[1.25pt]
 \multirow{2}{*}{\textbf{Methods}} & \textbf{\#Activated} & \multirow{2}{*}{\textbf{MMLU (5)}} & \multirow{2}{*}{\textbf{CMMLU (5)}} & \multirow{2}{*}{\textbf{NQ (32)}} & \multirow{2}{*}{\textbf{GSM8K(8)}} & \multirow{2}{*}{\textbf{TruthfulQA}} \\ 
 & \textbf{\ Heads (\%)} \\
 \midrule
 LLaMA3-8B-stage1  & 100 & 66.2 & 66.0 & 28.1 & 58.6 & 41.9 \\
 \cmidrule(rl){1-2} \cmidrule(rl){3-7}
 \rowcolor{aliceblue!60} \textbf{MoH-LLaMA3-8B} & 75 & 65.8 & 64.4 & 28.3 & 56.9 & 44.0 \\
 \midrule
 \midrule
 \multirow{2}{*}{\textbf{Methods}} & \textbf{\#Activated}  & \multirow{2}{*}{\textbf{HellaSwag (10)}} & \multirow{2}{*}{\textbf{LogiQA}} & \multirow{2}{*}{\textbf{BoolQ (32)}} & \multirow{2}{*}{\textbf{LAMBADA}} & \multirow{2}{*}{\textbf{SciQ}} \\ 
 & \textbf{\ Heads (\%)} \\
 \midrule
 LLaMA3-8B-stage1  & 100 & 79.4 & 30.4 & 85.1 & 75.8 & 92.2 \\
 \cmidrule(rl){1-2} \cmidrule(rl){3-7}
 \rowcolor{aliceblue!60} \textbf{MoH-LLaMA3-8B} & 75 & 80.1 & 30.3 & 84.0 & 76.4 & 92.2 \\
 \midrule
 \midrule
 \multirow{2}{*}{\textbf{Methods}} & \textbf{\#Activated}  & \multirow{2}{*}{\textbf{PIQA}} & \multirow{2}{*}{\textbf{WinoGrande}} & \multirow{2}{*}{\textbf{ARC-E}} & \multirow{2}{*}{\textbf{ARC-C (25)}}  & \multirow{2}{*}{\textbf{Average}} \\ 
 & \textbf{\ Heads (\%)} \\
 \midrule
 LLaMA3-8B-stage1  & 100 & 79.1 & 73.0 & 70.9 & 59.6 & 64.7 \\
 \cmidrule(rl){1-2} \cmidrule(rl){3-7}
 \rowcolor{aliceblue!60} \textbf{MoH-LLaMA3-8B} & 75 & 78.8 & 72.9 & 72.5 & 60.1 & \textbf{64.8} \\
\bottomrule[1.25pt]
\end{tabular}
}
\label{apdx tab: LLaMA3-8B-stage1}
\end{table*}

\begin{figure}[thp]
\centering
\vspace{-.4em}
\includegraphics[width=0.7\textwidth]{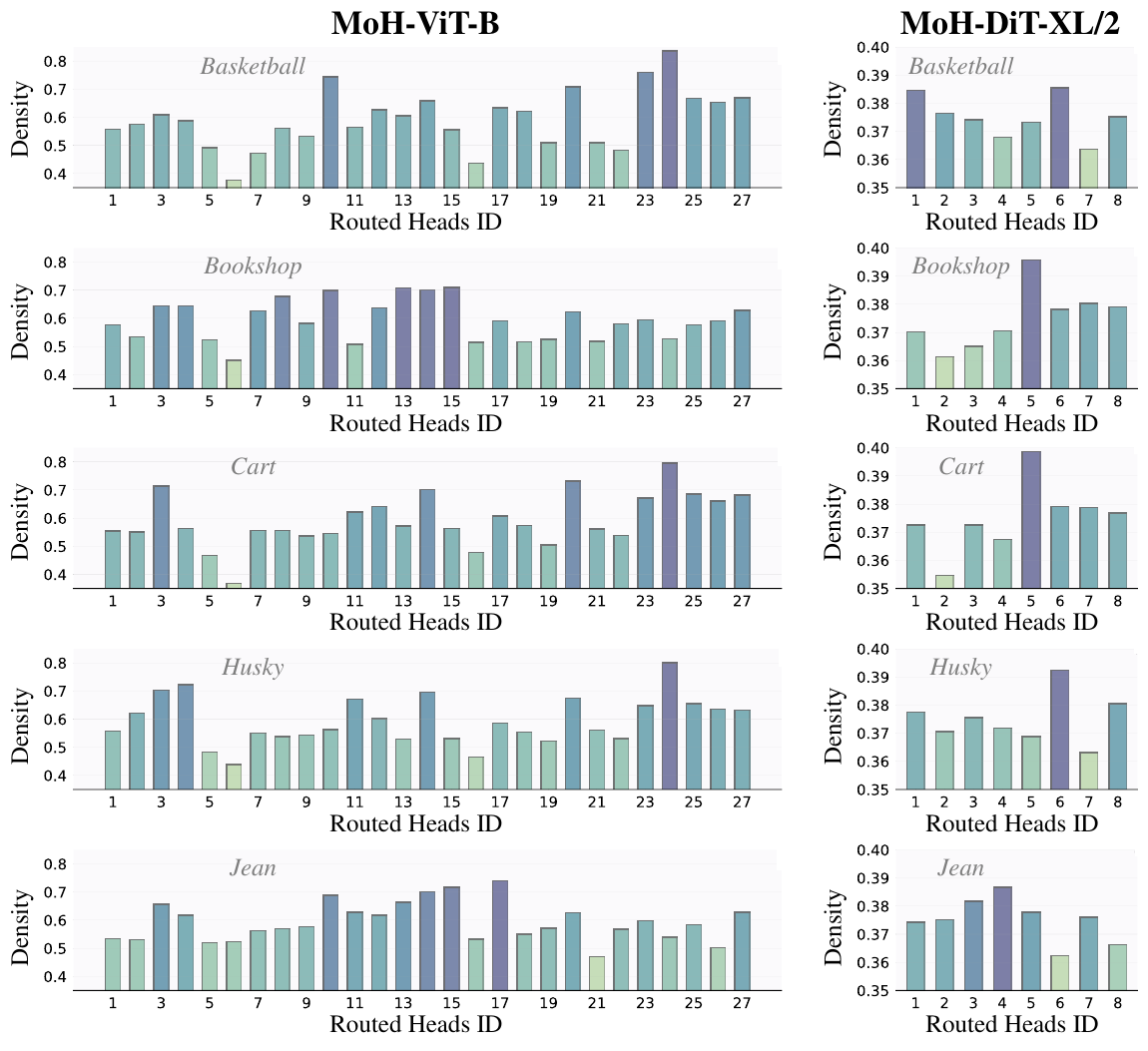}
\vspace{-1.4em}
\caption{\textbf{{Additional visualization of the head load distribution in the final MoH layer.}} {MoH-ViT-B activates 75\% of the attention heads. MoH-DiT-XL/2 activates 90\% of the attention heads.}}
\vspace{-1.em}
\label{apdx fig: fig0}
\end{figure}

\section{Additional Experiments}\label{appendix:Additional Experiments}
\myparagraph{{Comparison between MoH-LLaMA3-8B and LLaMA3-8B-stage1.}} 
{We divide the training process into two stages. Tab.~\ref{apdx tab: LLaMA3-8B-stage1} shows the comparison between MoH-LLaMA3-8B and the model at the end of the first training stage (LLaMA3-8B-stage1). As shown in Tab.~\ref{apdx tab: LLaMA3-8B-stage1}, MoH-LLaMA3-8B quickly recovers the performance of LLaMA3-8B-stage1 within a training budget of 100B tokens. Notably, in English language tasks, MoH-LLaMA3-8B surpasses LLaMA3-8B-stage1 while using only 75\% of the attention heads. However, for Chinese language and math tasks, the recovery performance of the MoH model is not as strong as for English. For example, MoH-LLaMA3-8B achieves an accuracy of 64.4\% on CMMLU, compared to 66.0\% for LLaMA3-8B-stage1. We attribute this to the fact that the model's Chinese and mathematical capabilities are primarily established during the first training stage. Since the first training stage uses only 300B tokens, significantly less than the 15T tokens in LLaMA3-8B's pre-training, the model's abilities in these areas are not fully stable. In the second training stage, after switching to the MoH model, the model experiences more significant forgetting in Chinese and math tasks. Overall, as shown in Tab.~\ref{apdx tab: LLaMA3-8B-stage1}, MoH-LLaMA3-8B achieves an average accuracy of 64.8\% across 14 benchmarks, outperforming LLaMA3-8B-stage1 by utilizing only 75\% of the attention heads.}

\myparagraph{{Effect of the Activated Head Ratio.}} 
{As shown in Tab.~\ref{apdx tab: ablation2}, activating more attention heads generally leads to improved model performance. These results are intuitive, as activating more attention heads equates to utilizing more parameters and performing additional computations on the input.}

\begin{table*}[ht]
\footnotesize
\centering
\caption{\textbf{{Ablation study on the impact of the activated head ratio.}} {All results are from MoH-ViT-S, by using a training budget of 100 epochs.}}
\vspace{.5em}
{
\begin{tabular}{lccccccccc}
\toprule[1.25pt]
 {\textbf{Activated Heads}} & 50\% & 55\% & 60\% & 65\% & 70\% & 75\% & 80\% \\ 
 \midrule
 {\textbf{Accuracy (\%)}} & 78.32 & 78.38 & 78.44 & 78.50 & 78.42 & 78.58 & \textbf{78.78} \\
\bottomrule[1.25pt]
\end{tabular}
}
\vspace{-1.em}
\label{apdx tab: ablation2}
\end{table*}

\section{Additional Qualitative Analysis}\label{appendix:Additional Qualitative Analysis}
\myparagraph{{Additional Visualization of the Head Load Distribution.}}
{We provide additional visualization of the head load distribution in Fig.~\ref{apdx fig: fig0}. As illustrated in both Fig.~\ref{fig3} and Fig.~\ref{apdx fig: fig0}, there is notable variation in attention head assignments across different categories and task topics. This suggests that the MoH model adapts to a wide range of tasks by utilizing distinct head assignment patterns. This ability enables MoH to allocate attention heads more effectively to specific task types, leading to more efficient parameter utilization compared to standard multi-head attention.}

\begin{figure}[thp]
\centering
\includegraphics[width=0.7\textwidth]{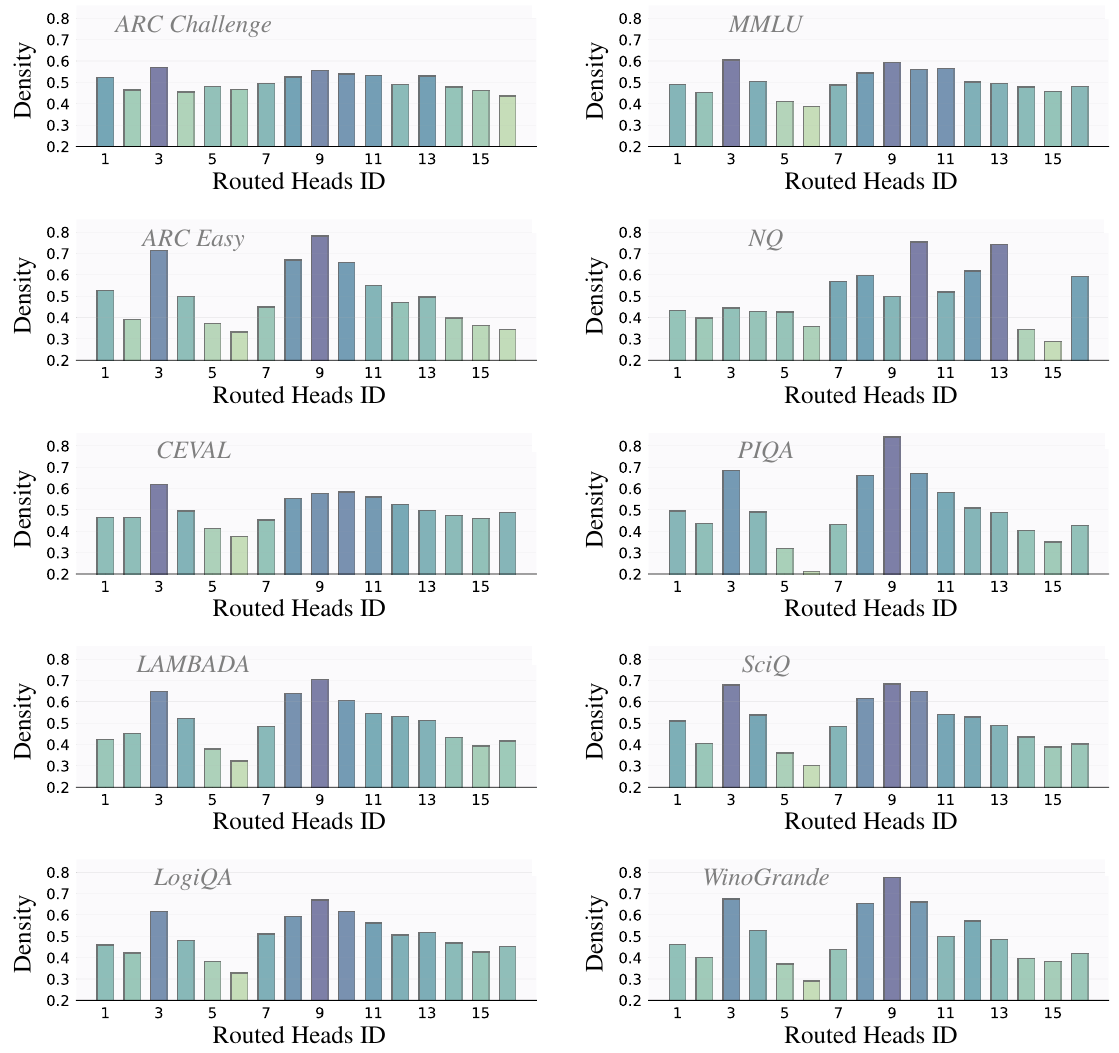}
\vspace{-1.em}
\caption{\textbf{{Additional visualization of the head load distribution in MoH-LLaMA3-8B.}}}
\vspace{-1.em}
\label{apdx fig: fig1}
\end{figure}

\myparagraph{{Additional Visualization of the Head Load Distribution in MoH-LLaMA3-8B.}}
{We provide additional visualization of the head load distribution in Fig.~\ref{apdx fig: fig1}. As shown in Fig.~\ref{apdx fig: fig1}, MoH-LLaMA3-8B exhibits similar characteristics to MoH-LLMs trained from scratch, with significant variation in attention head assignments across different categories and task topics. This indicates that continue-tuning enables the model to adopt different head assignment patterns quickly. These results demonstrate that pre-trained multi-head attention models can be effectively continue-tuned into MoH models, significantly broadening the applicability of the proposed MoH approach.}

\myparagraph{{Additional Visualization of the Head Routing Score Distribution.}}
{We provide additional visualization of the head routing score distribution in Fig.~\ref{apdx fig: fig1 vit}, Fig.~\ref{apdx fig: fig1 dit}, and Fig.~\ref{apdx fig: fig1 llm}. As illustrated in Fig.~\ref{apdx fig: fig1 vit}, Fig.~\ref{apdx fig: fig1 dit}, and Fig.~\ref{apdx fig: fig1 llm}, these head routing scores also vary across categories and task types. This dynamic weighting mechanism allows MoH to adjust the importance of each head in response to different task requirements, further enhancing its flexibility and performance. Besides, we find that the routing scores of shared heads change more across categories than those of routing headers. We consider this because routed heads adapt to different categories by adjusting their activation, while shared heads remain activated all the time. Therefore, shared heads primarily rely on changes in routing scores to adapt to different categories.}

\myparagraph{Images Generated from the Proposed MoH-DiT-XL/2 Model.}
Fig.~\ref{apdx fig: fig2} shows samples generated by our class-conditional MoH-DiT-XL/2 model. These results demonstrate the ability of MoH-DiT-XL/2 to generate semantically correct content with accurate spatial relationships.

\section{Details of Quantitative Evaluations for LLMs}\label{appendix:Details of Quantitative Evaluations}
\footnotetext[4]{\href{https://github.com/EleutherAI/lm-evaluation-harness}{https://github.com/EleutherAI/lm-evaluation-harness}}

We conduct comparative comparisons of MoH-LLM (MoH-LLaMA3-8B) against vanilla LLMs~(LLaMA3-8B). The evaluation is performed on multiple key benchmarks using the Eleuther AI Language Model Evaluation Harness\footnotemark[4]~\citep{eval-harness}, a unified framework for testing generative language models across a wide range of tasks. The benchmarks used for evaluation include:

\begin{figure}[tp]
\centering
\includegraphics[width=0.7\textwidth]{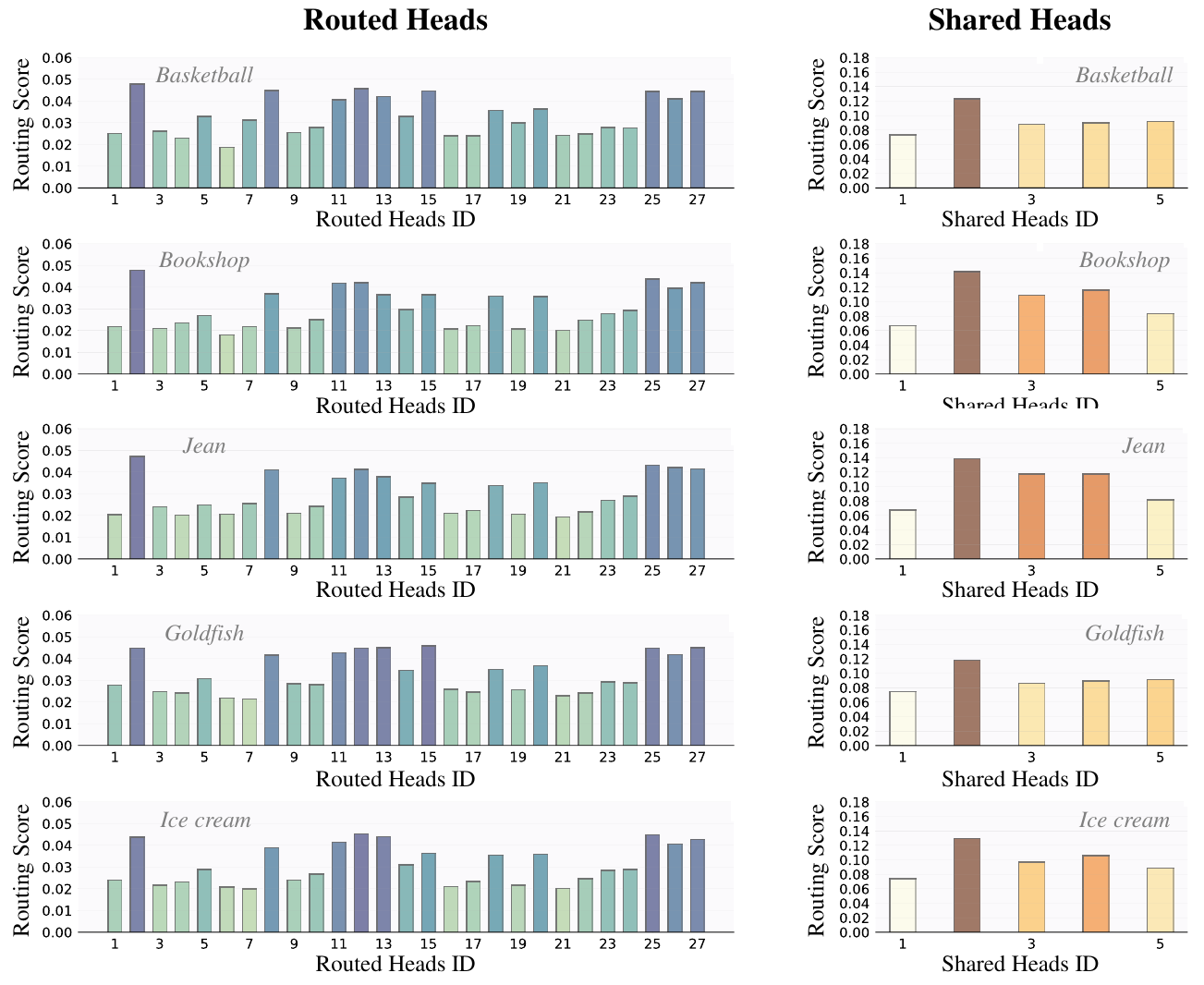}
\vspace{-1.em}
\caption{\textbf{{Additional visualization of the head routing score distribution in MoH-ViT-B.}} {MoH-ViT-B activates 75\% of the attention heads.}}
\vspace{-1.em}
\label{apdx fig: fig1 vit}
\end{figure}

\textbf{ARC}~\citep{clark2018think} is a multiple-choice question-answering resource featuring questions from science exams for grades 3 to 9. It is divided into two partitions: Easy and Challenge, with the latter containing more difficult questions that necessitate reasoning. Most questions offer four answer choices, while less than 1\% feature either three or five choices. Additionally, ARC includes a supporting knowledge base with 14.3 million unstructured text passages. We report 0-shot accuracy on ARC Easy and 25-shot accuracy on ARC Challenge.

\textbf{LAMBADA}~\citep{paperno2016lambada} is an open-ended cloze task consisting of approximately 10,000 passages from BooksCorpus, where the objective is to predict a missing target word in the last sentence of each passage. The missing word is always the last word of the final sentence, with no options provided. We report 0-shot accuracy on LAMBADA.

\textbf{LogiQA}~\citep{liu2020logiqa} comprises 8,678 question-and-answer instances that encompass various types of deductive reasoning. The dataset serves as a benchmark for reexamining logical AI within the context of deep learning in NLP. We report 0-shot accuracy on LogiQA.

\begin{figure}[tp]
\centering
\vspace{-.4em}
\includegraphics[width=0.7\textwidth]{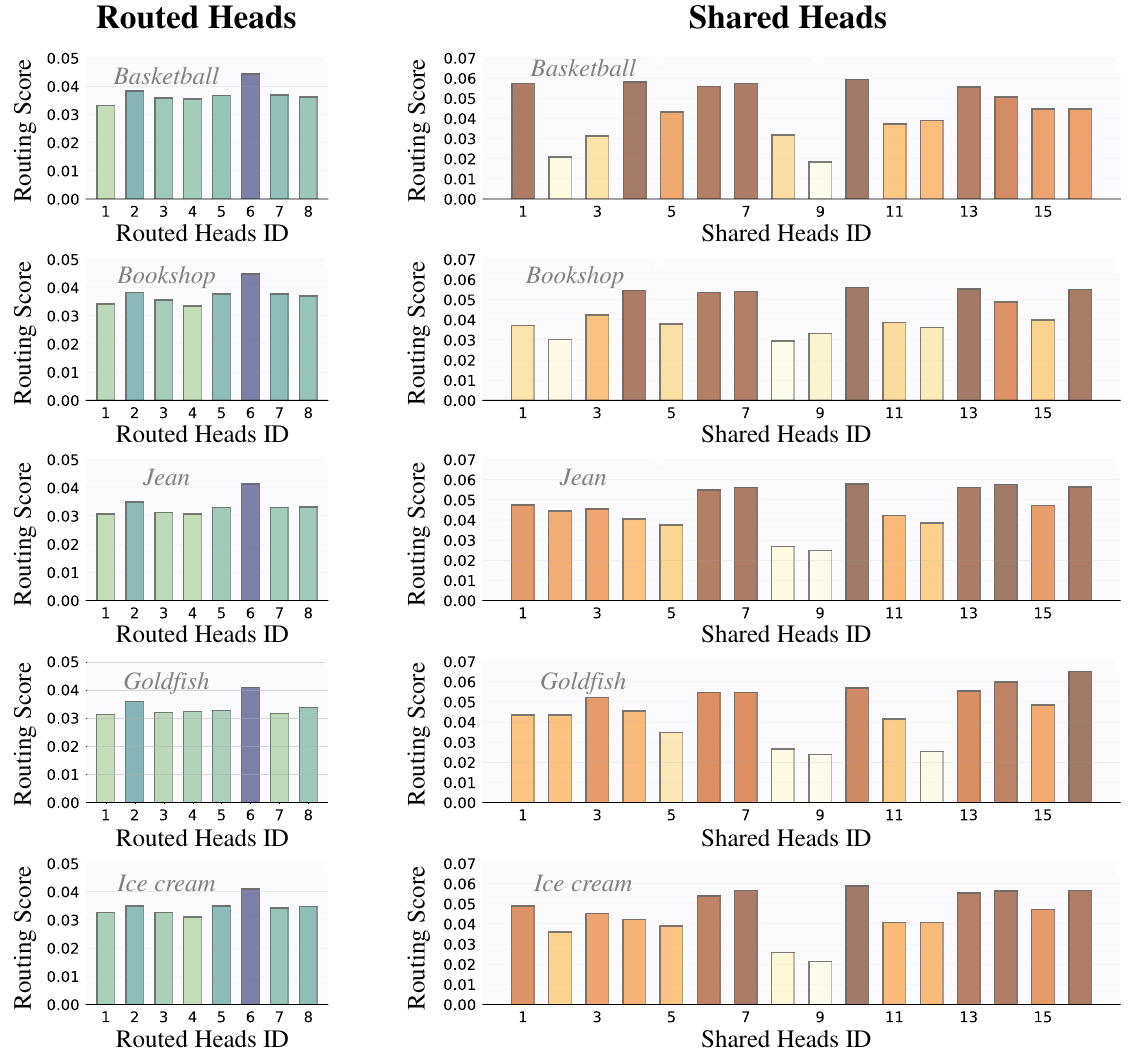}
\vspace{-1.2em}
\caption{\textbf{{Additional visualization of the head routing score distribution in MoH-DiT-XL/2.}} {MoH-DiT-XL/2 activates 90\% of the attention heads.}}
\vspace{-1.2em}
\label{apdx fig: fig1 dit}
\end{figure}

\textbf{PIQA}~\citep{bisk2020piqa} is a dataset designed for commonsense reasoning, aimed at evaluating the physical knowledge of current models. We report 0-shot accuracy on PIQA.

\textbf{SciQ}~\citep{welbl2017crowdsourcing} includes 13,679 crowdsourced science exam questions covering subjects such as Physics, Chemistry, and Biology. Each question is presented in a multiple-choice format with four answer options, and for most questions, an additional paragraph provides supporting evidence for the correct answer. We report 0-shot accuracy on SciQ.

\textbf{WinoGrande}~\citep{sakaguchi2021winogrande} is a large-scale dataset comprising 44,000 problems, inspired by the original WSC design but enhanced to increase both its scale and difficulty. We report 0-shot accuracy on WinoGrande.

\textbf{HellaSwag}~\citep{zellers2019hellaswag} is a challenging dataset designed to evaluate commonsense natural language inference, which proves difficult for state-of-the-art models but poses no significant challenge for humans. We report the accuracy for the 10-shot HellaSwag.

\textbf{MMLU}~\citep{hendrycks2020measuring} is a benchmark designed to assess models' knowledge acquired during pretraining, making it more challenging and human-like in evaluation. It covers 57 subjects across STEM, humanities, social sciences, and more, ranging from elementary to advanced professional levels. The benchmark tests both world knowledge and problem-solving skills, with subjects spanning traditional areas like math and history to specialized fields such as law and ethics, offering a comprehensive tool for identifying model blind spots. We report the accuracy for the 5-shot MMLU.

\textbf{Natural Questions (NQ)}~\citep{kwiatkowski2019natural} is a question-answering dataset based on real, anonymized Google queries. Annotators label long and short answers (or null if no answer is found) from Wikipedia pages in the top 5 search results. The dataset includes 307,373 training examples, 7,830 development examples, and 7,842 test examples with 5-way annotations. We report the exact match score for 32-shot Natural Questions to measure the factual knowledge in the model.

\textbf{BoolQ}~\citep{clark2019boolq} is a question-answering dataset consisting of 15,942 yes/no questions. These questions are naturally occurring, and generated in unprompted and unconstrained contexts. Each example is provided as a triplet of (question, passage, and answer), with the page title optionally included as additional context. We report the accuracy for the 32-shot BoolQ.

\textbf{OpenbookQA}~\citep{OpenBookQA2018} is a question-answering dataset designed to assess understanding of elementary-level science, similar to open-book exams. It contains 5,957 multiple-choice questions based on a ``book'' of 1,326 core science facts. The dataset requires not only knowledge of these facts but also the application of broad common knowledge. It includes mappings from each question to the core fact it targets and additional common knowledge facts. The dataset also provides scores of human accuracy and clarity, as well as crowd-sourced data for further analysis. We report 0-shot accuracy on OpenbookQA.

\textbf{TruthfulQA}~\citep{lin-etal-2022-truthfulqa} is a benchmark designed to evaluate the truthfulness of a language model's responses. It consists of 817 questions across 38 categories, such as health, law, finance, and politics. The questions are crafted to reflect common false beliefs or misconceptions that might lead humans to answer inaccurately. We report 0-shot accuracy on TruthfulQA.

\begin{figure}[tp]
\centering
\includegraphics[width=0.7\textwidth]{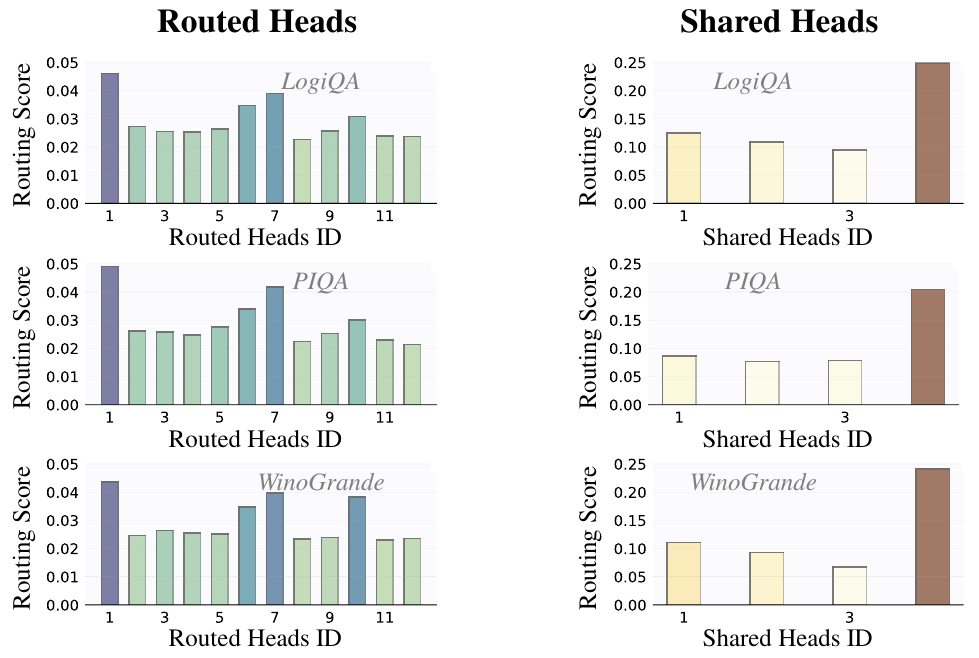}
\vspace{-1.em}
\caption{\textbf{{Additional visualization of the head routing score distribution in MoH-LLM-B.}} {MoH-LLM-B activate 75\% of the attention heads.}}
\label{apdx fig: fig1 llm}
\end{figure}

\textbf{GSM8K}~\citep{cobbe2021training} is a dataset containing 8.5K high-quality, linguistically diverse grade school math word problems. It is divided into 7.5K training problems and 1K test problems. Each problem requires 2 to 8 steps to solve, typically involving a sequence of elementary calculations using basic arithmetic operations. A capable middle school student should be able to solve all the problems, making the dataset suitable for evaluating multi-step mathematical reasoning. We report the exact match score for 8-shot GSM8K.

\textbf{CEVAL}~\citep{huang2024c} is a comprehensive Chinese evaluation suite designed to assess the advanced knowledge and reasoning abilities of LLMs in a Chinese context. It includes multiple-choice questions across four difficulty levels (middle school, high school, college, and professional) and spans 52 diverse disciplines. We report the accuracy for the 5-shot CEVAL.

\textbf{CMMLU}~\citep{li2023cmmlu} is a comprehensive Chinese benchmark designed to evaluate the knowledge and reasoning abilities of LLMs across various subjects, including natural sciences, social sciences, engineering, and humanities. We report the accuracy for the 5-shot CMMLU.

\begin{figure}[tp]
\centering
\includegraphics[width=1\textwidth]{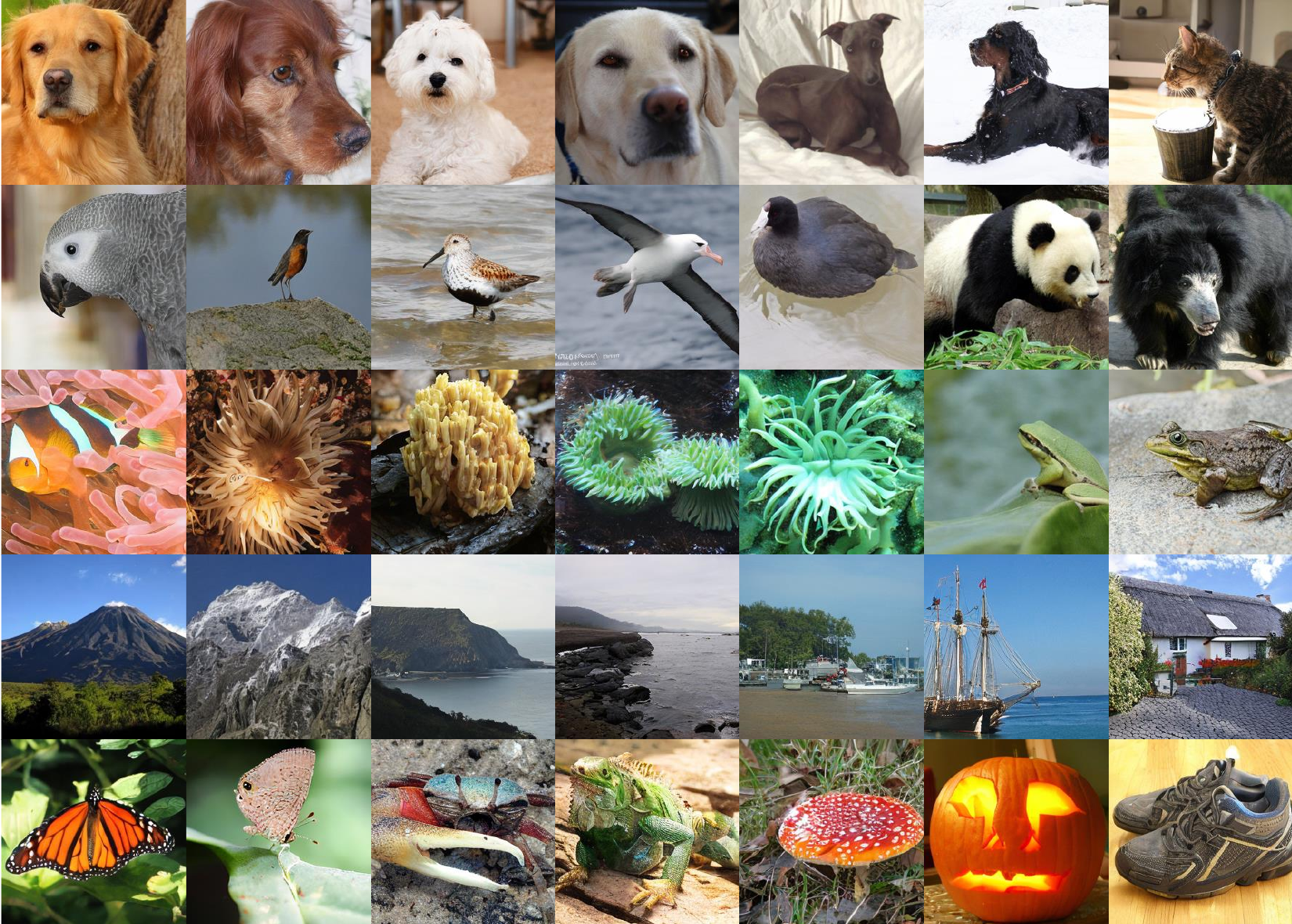}
\caption{\textbf{Images generated from the proposed MoH-DiT-XL/2 model.} We show samples generated from our class-conditional MoH-DiT-XL/2 model trained on ImageNet at 256$\times$256 resolution. MoH-DiT-XL/2 activates 90\% of the attention heads.}
\label{apdx fig: fig2}
\end{figure}


\end{document}